%% file: manuscript.tex
\documentclass[twocolumn]{svjour3}          
\smartqed  
\usepackage{graphicx}
\usepackage[numbers]{natbib}
\usepackage{times}
\usepackage{epsfig}
\usepackage{amsmath}
\usepackage{amssymb}
\usepackage{lineno}

\usepackage[pagebackref=true,breaklinks=true,letterpaper=true,colorlinks,bookmarks=false,citecolor=blue,linkcolor=blue]{hyperref}

\usepackage{booktabs}
\usepackage{amsmath}
\usepackage{amssymb}
\usepackage{verbatim}
\usepackage{multirow, makecell}
\usepackage{lineno}
\usepackage{enumerate}
\usepackage{amsfonts}
\usepackage{gensymb}
\usepackage{bm}
\usepackage{bbding}
\usepackage{comment}
\usepackage{color}
\usepackage{diagbox}
\usepackage{tabularx}
\usepackage{rotating}
\newcolumntype{L}[1]{>{\raggedright\let\newline\\\arraybackslash\hspace{0pt}}m{#1}}
\newcolumntype{C}[1]{>{\centering\arraybackslash}m{#1}}
\newcolumntype{R}[1]{>{\raggedleft\let\newline\\\arraybackslash\hspace{0pt}}m{#1}}
\newlength\savewidth

\usepackage{colortbl}
\definecolor{Gray}{gray}{0.9}
\definecolor{GrayT}{gray}{0.4}

\newcommand{\eg}{{\em e.g.}}
\newcommand{\ie}{{\em i.e.}}

\usepackage{subcaption}
\usepackage[labelfont=bf]{caption}
\definecolor{light-blue}{RGB}{186,175,255}
\definecolor{light-red}{RGB}{255,137,165}
\usepackage{arydshln} 
\usepackage{booktabs}
\usepackage{multirow}
\usepackage{amsmath}
\usepackage{etoolbox}
\usepackage{bm}
\usepackage{mathtools}
\usepackage{algorithm}
\usepackage{algpseudocode}
\usepackage{xspace}

\usepackage{multirow, makecell}
\usepackage{colortbl}  
\usepackage{xcolor}
\usepackage{array}
\definecolor{Gray}{gray}{0.94}
\definecolor{liGray}{gray}{0.5}
\definecolor{LightCyan}{rgb}{0.88,1,1}
\usepackage{mathrsfs}

\usepackage{utfsym}
\begin{document}
\begin{sloppypar}

\title{TryOn-Adapter: Efficient Fine-Grained Clothing Identity Adaptation for High-Fidelity Virtual Try-On
}


\author{Jiazheng Xing      \and
        Chao Xu    \and
        Yijie Qian       \and
        Yang Liu       \and
         Guang Dai  \and \\
    Baigui Sun \and
     Yong Liu \and
      Jingdong Wang
}


\institute{
Jiazheng Xing \and Yijie Qian \and Yong Liu
\at
        Laboratory of Advanced Perception on Robotics and Intelligent Learning, College of Control Science and Engineering, Zhejiang University, Hangzhou 310027, Zhejiang, China  \\
              \email {\{jiazhengxing, yijieqian\}@zju.edu.cn  and yongliu@iipc.zju.edu.cn} 
           \and
           Chao Xu \and Yang Liu \and  Baigui Sun \at
              Alibaba Group \\
              \email{\{xc264362, ly261666, baigui.sbg\}@alibaba-inc.com }
              \and
          Guang Dai \at
          SGIT AI Lab, State Grid Shaanxi Electric Power Company \\
          \email {guang.gdai@gmail.com}
          \and
          Jingdong Wang \at 
          Baidu Inc. \\
          \email {wangjingdong@baidu.com}
    }

\date{Received: date / Accepted: date}
\maketitle


\input{sec/0_abstract} 
 \begin{figure*} [h]
		\centering
		\includegraphics[width=0.9\linewidth ]{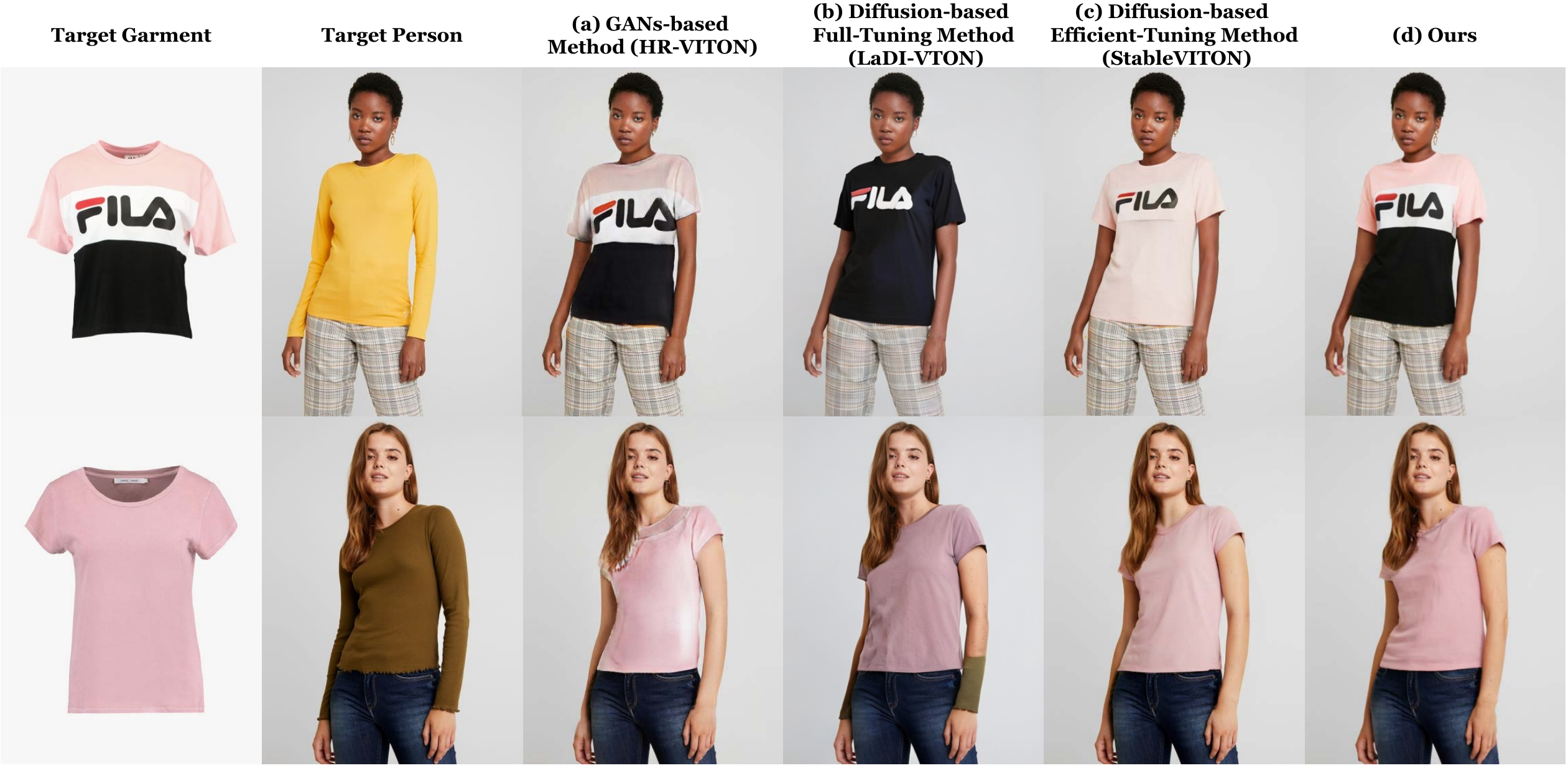}
  		\vspace{-10pt}
	\caption{Performance comparison of three different methods on VITON-HD dataset at 512 $\times$ 384 resolution, including our TryOn-Adapter, GANs-based method HR-VITON~\cite{rombach2022high}, Diffusion-based method LaDI-VTON~\cite{morelli2023ladi} and StableVITON~\cite{kim2023stableviton}. Our method generates high-quality results and exhibits strong clothing identity preservation capability, \ie, consistent color style and logo textures, as well as a smooth transition between long and short sleeves.}  
		\label{fig:intro}
		\vspace{-13pt}
\end{figure*}
\input{sec/1_intro}

\input{sec/2_related_work}

\input{sec/3_method}
\input{sec/4_exp}
\input{sec/5_conclusion}

{\small
\bibliographystyle{spbasic}
\bibliography{egbib}
}
\end{sloppypar}
\end{document}

%% file: sec/0_abstract.tex
\begin{abstract}

Virtual try-on focuses on adjusting the given clothes to fit a specific person seamlessly while avoiding any distortion of the patterns and textures of the garment. However, the clothing identity uncontrollability and training inefficiency of existing diffusion-based methods, which struggle to maintain the identity even with full parameter training, are significant limitations that hinder the widespread applications. In this work, we propose an effective and efficient framework, termed TryOn-Adapter. Specifically, we first decouple clothing identity into fine-grained factors: style for color and category information, texture for high-frequency details, and structure for smooth spatial adaptive transformation. Our approach utilizes a pre-trained exemplar-based diffusion model as the fundamental network, whose parameters are frozen except for the attention layers. We then customize three lightweight modules (Style Preserving, Texture Highlighting, and Structure Adapting) incorporated with fine-tuning techniques to enable precise and efficient identity control. Meanwhile, we introduce the training-free T-RePaint strategy to further enhance clothing identity preservation while maintaining the realistic try-on effect during the inference. 
Our experiments demonstrate that our approach achieves state-of-the-art performance on two widely-used benchmarks. Additionally,
compared with recent full-tuning diffusion-based methods, we only use about half of their tunable parameters during training. The code will be made publicly available at  \href{https://github.com/jiazheng-xing/TryOn-Adapter}{https://github.com/jiazheng-xing/TryOn-Adapter}.

\keywords{Virtual Try-On  \and Large-Scale Generative Models \and  Diffusion Model   \and Identity Preservation}

\end{abstract}
\vspace{-10pt}


%% file: sec/1_intro.tex
\section{Introduction}
\label{sec:intro}

Virtual try-on~\cite{han2018viton, wang2018toward, choi2021viton, morelli2022dress} aims to enable users to naturally try on new category clothes in the target regions by giving an image of the garment and an image of the person while preserving the non-target regions. The core of this task lies in maintaining the pattern and texture of the clothes, termed \textit{clothing identity}, unchanged in various conditions.
Considering the scarcity of high-quality paired datasets, the current works usually follow a two-stage design~\cite{choi2021viton, morelli2022dress, lee2022high, xie2023gp, gou2023taming}: target garment deformation and composite generation. The former~\cite{ge2021parser, han2019clothflow, he2022style} focuses on transferring the original clothing into the desired form based on the posture and body shape of the given person. Despite providing a prior warped template, direct blending produces severe artifacts when encountering occlusion and large shape differences. Therefore, the latter is introduced for further refinement with a powerful generative model. Concretely, most of the previous works~\cite{ge2021parser, bai2022single, lee2022high, morelli2022dress} have relied on the GANs~\cite{goodfellow2014generative}, but they suffer from unstable training~\cite{gulrajani2017improved} and mode collapse~\cite{miyato2018spectral}, leading to the detail loss in their generated results, especially for the highly patterned garments, as shown in Fig.~\ref{fig:intro} column 3. More recently, diffusion models~\cite{song2020denoising, ho2020denoising, rombach2022high} have attracted widespread attention and permeated into the virtual try-on. Recently, two diffusion-based models have regarded try-on as an inpainting task. DCI-VTON~\cite{gou2023taming} is built upon the exemplar-based image generation method, harnessing its ability to preserve irrelevant areas while focusing on fusing the given garment into the target area. LaDI-VTON~\cite{morelli2023ladi} employs the textual-inversion technique to fill the target areas. However, they cannot achieve satisfactory results due to insufficient exploration of identity-preserving modules, even tuning all parameters of UNet for adaptive learning. As shown in Fig.~\ref{fig:intro} column 4, the color and textures of their generated clothes are completely different from the target clothes (row 1), and the transition from long sleeves to short sleeves exhibits obvious artifacts (row 2).

Although diffusion-based garment composition generations have progressed, they lack in-depth thinking in two key aspects. 
(1) \textit{Identity controllability}. Previous methods~\cite{yang2023paint,gou2023taming} utilize the class token of CLIP embeddings obtained from the reference garment image. However, this global vectorized feature, when directly integrated into the UNet, fails to retain identity cues. By contrast, this work decouples the garment characteristics into three fine-grained factors to simplify identity preservation, \ie, style (color and category information), texture (high-frequency details such as patterns, logo, and text), and structure (smooth transition when under different pose or body shape, as well as a significant difference between the original and target clothing, such as the aforementioned long and short sleeves issues). 
(2) \textit{Training efficiency}. Diffusion-based methods usually suffer from low training efficiency, especially in a fully fine-tuned manner. To tackle this problem, Parameter-Efficient Fine-Tuning techniques (PEFT), such as ControlNet~\cite{yang2023paint}, T2I-Adapter~\cite{mou2023t2i}, and GLIGEN~\cite{li2023gligen}, employing a small number of training parameters to control the denoising process. {It is worthwhile to consider how to introduce efficient training modules or even training-free mechanisms into the try-on task without sacrificing performance.} Notably, concurrent work StableVITON~\cite{kim2023stableviton} circumvents full-tuning, yet their tenuous representation of clothing identity (only a single image) results in continued difficulty in producing satisfactory outputs. As shown in Fig.~\ref{fig:intro} column 5, they exhibit inconsistent color style and logo textures.



{To realize more identity-controllable and training-efficient virtual try-on, we propose a novel paradigm, termed \textbf{TryOn-Adapter}, which follows the Paint-by-Example framework and customizes three lightweight components according to the decoupled factors to effectively control hierarchical identity cues.}
Specifically, we start by freezing all the parameters in the UNet blocks except for the attention layers, which transfer the universal pre-trained model to the specific try-on task with minimum trainable parameters. For style preservation, we utilize both patch and class tokens to learn comprehensive style representation, with the former compensating for the lack of detailed identity in the latter. Furthermore, due to the limitation of CLIP in capturing the complex color style, we further enhance the patch tokens with visual features embedded in the VAE encoder through an adaptive transfer module. To avoid disturbing the feature distribution of the pre-trained model, inspired by GLIGEN~\cite{li2023gligen}, we insert trainable gated self-attention layers in all layers to inject the updated patch tokens into the frozen backbone. Moreover, to preserve the texture, a post-processed high-frequency feature map is incorporated as a texture refinement guidance to highlight the local details. For another factor involving spatial cues, we take the segmentation map, {obtained by a rule-based training-free extraction method,} as the structure condition to explicitly rearrange the target areas of the body and clothing to conform to the warped cloth. We follow the T2I-Adapter~\cite{mou2023t2i} to inject the above two conditions into UNet by two lightweight networks {incorporated a well-designed position attention module that helps amplify the spatial cues.}
 {During the inference phase, we introduce a time-partially function on the training-free technique RePaint~\cite{lugmayr2022repaint, avrahami2023blended}, termed T-RePaint, to further enhance the clothing identity without compromising the overall image fidelity. 
Additionally, a learnable latent blending module is integrated within the autoencoder to produce more visually consistent results.}


In this way, we preserve the hierarchical identity details of the given garment without full fine-tuning, as illustrated in Fig.~\ref{fig:intro} column 6.



In summary, we present the following contributions.
\begin{itemize}
\item We propose a novel, effective, and efficient framework for virtual try-on, \textbf{TryOn-Adapter}, to maintain the identity of the given garment with low consumption.

\item We decouple clothing identity into fine-grained factors: style, texture, and structure, represented by the global class token and enhanced patch token embeddings, high-frequency feature map, and segmentation maps, respectively. Each factor incorporates a tailored lightweight module and injection mechanism to achieve precise and efficient identity control. {Meanwhile, we introduce a training-free technique, T-RePaint, to further reinforce the clothing identity preservation while maintaining the realistic try-on effect during the inference.}


\item Extensive experiments on two widely used datasets have shown that our method can achieve outstanding performance with minor trainable parameters.
\end{itemize}
\vspace{-10pt}

%% file: sec/2_related_work.tex
\section{Related Work}
\label{sec:formatting}
\subsection{Image-based Virtual Try-On}
To avoid distortion of garment image textures and confusion of the identities as much as possible, the image-level virtual try-on~\cite{yang2020towards, issenhuth2020not, han2018viton, wang2018toward, choi2021viton, morelli2022dress, chen2023size,li2023virtual}  task is typically divided into two stages: the target garment deformation stage and the composite generation stage. For the first stage, the Thin Plain Spine (TPS) method was commonly employed to deform clothing in previous works~\cite{han2018viton, ge2021disentangled, minar2020cp, zheng2019virtually, yang2020towards}, which is limited to offering only basic deformation processing. Furthermore, many flow-based works~\cite{ge2021parser, han2019clothflow, he2022style, bai2022single} have been proposed, aiming to build the appearance flow field between clothing and corresponding regions of the human body to better deform the clothing for a more natural fit to the body. In our work, we adopt the flow-based method PF-AFN~\cite{ge2021parser} to accomplish the rough deformation of the clothing in the first stage. The second stage can be classified into two categories: the GANs-based methods and the Diffusion-based methods. GANs-based methods~\cite{ge2021parser, bai2022single, lee2022high, morelli2022dress,lewis2021tryongan} inherit the weaknesses of Generative Adversarial Networks (GANs)~\cite{goodfellow2014generative}, such as unstable training~\cite{gulrajani2017improved} and mode drop in the output distribution~\cite{miyato2018spectral}, leading to the problem of detail loss in their generated results. Specifically, FashionGAN~\cite{zhu2017your} generates the image conditioned on textual descriptions and semantic layouts, TryOnGAN~\cite{lewis2021tryongan} trains a pose-conditioned StyleGAN2~\cite{karras2020analyzing} on unpaired fashion images, and so on. Unlike the former, Diffusion-based methods~\cite{yang2023paint, gou2023taming, morelli2023ladi, baldrati2023multimodal, li2023warpdiffusion} with a more stable training procedure can provide superior image generation quality, including distributional coverage and flexibility. {Specifically, MGD~\cite{baldrati2023multimodal} is the first latent diffusion model defined for humancentric fashion image editing, conditioned by multimodal inputs such as text, body pose, and sketches. LaDI-VTON~\cite{morelli2023ladi} exploits the textual inversion technique for the first time in this task, demonstrating its capability in conditioning the generation process. DCI-VTON~\cite{gou2023taming} treats the virtual try-on task as an inpainting task and adds the warped clothes to the input of the diffusion model as the local condition. However, they encounter issues with high resource consumption and difficulties in controlling their clothing identity. StableVITON~\cite{kim2023stableviton}, a more recent work, only trains the proposed zero cross-attention blocks and SD encoder, but they still directly use the given garment image to provide clothing cues, which makes it difficult for the network to capture details. By contrast, our method decouples the complex clothing into fine-grained features and tailors them with carefully chosen fine-tuning techniques to significantly enhance the preservation of the given garment without incurring excessive training consumption.
}

\subsection{Diffusion Models}
Recently, the Denoising Diffusion Probabilistic Model (DDPM)~\cite{ho2020denoising, sohl2015deep} has emerged as a critical technology in image synthesis, renowned for its ability to generate high-fidelity images from a normal distribution by reversing the noise addition process. In response to the computational complexity and resource requirements of DDPM, the Latent Diffusion Model~\cite{rombach2022high} (LDM)  efficiently performs diffusion and denoising in the latent space through its optimized encoder-decoder architecture, streamlining the generation process. Based on the unique advantages shown by the Diffusion model in preserving image details, many works in text2image generation~\cite{gal2022image, dhariwal2021diffusion, ho2022classifier, wei2023elite, ramesh2022hierarchical,li2023gligen}, image editing~\cite{mou2023t2i,zhang2023adding, saharia2022photorealistic,nichol2021glide}, and subject-driven generation~\cite{chen2023anydoor,bhunia2023person,yang2023paint,shi2023instantbooth,wang2022hs} have recently emerged. The success of these 
previous works have provided ample inspiration for image-based virtual Try-On.

\vspace{-5pt}

%% file: sec/3_method.tex
\section{Method}

\subsection{Architecture Overview}
 \begin{figure*} [h]
		\centering
		\includegraphics[width=\linewidth ]{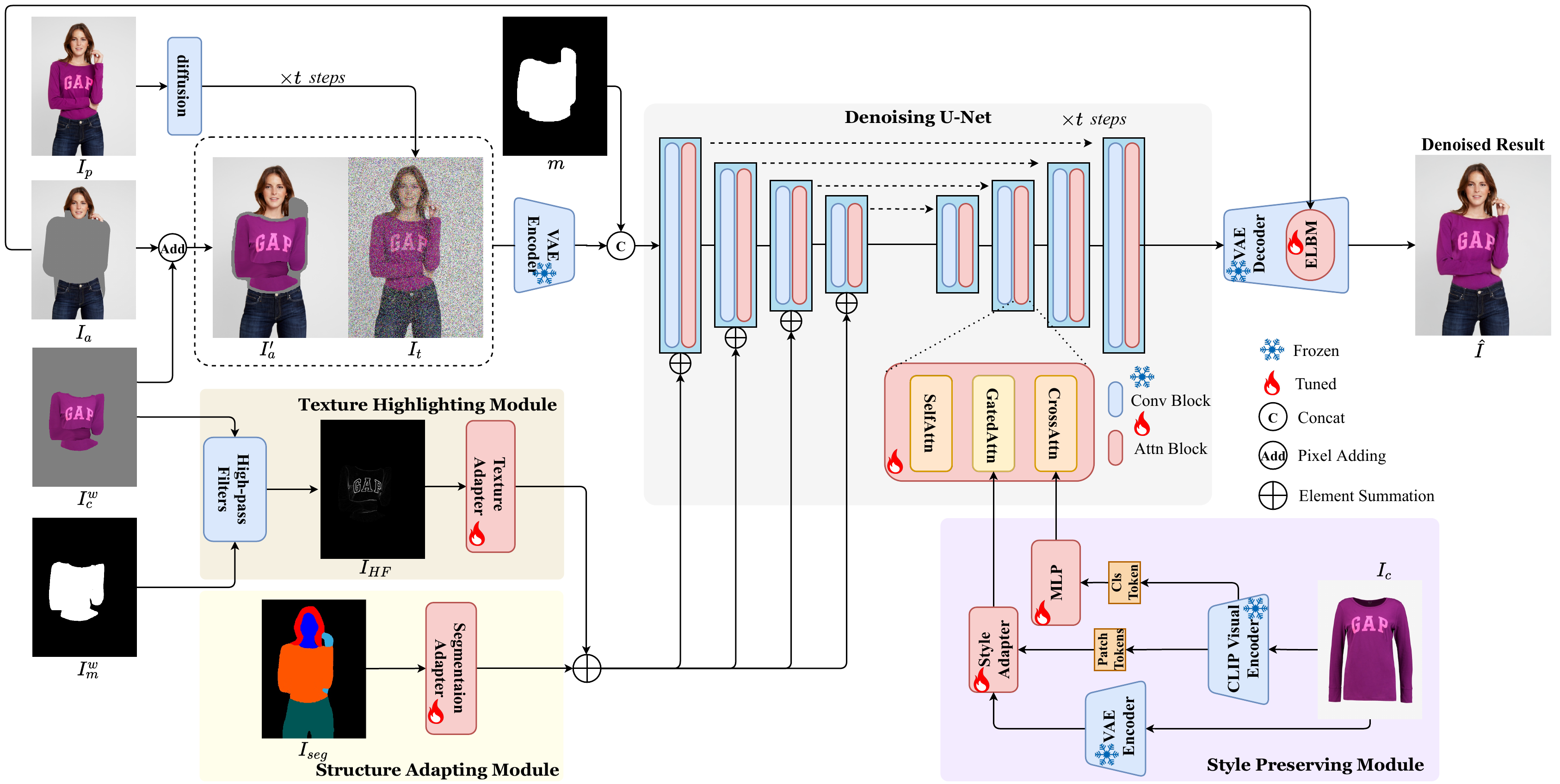}
  		\vspace{-15pt}
	\caption{The overall architecture of our \textbf{TryOn-Adapter} is composed of {five} parts: 1) the pre-trained stable diffusion model with fixed parameters except for attention layers; 2) the Style Preserving module aimed to preserve the overall style of the garment, including color and category information; 3) the Texture Highlighting module focuses on refining the high-frequency details. 4) the Structure Adapting module compensates for unnatural areas caused by clothing changes. {5) the Enhanced Latent Blending Module focuses on consistent visual quality.}}  
		\label{fig:main}
		\vspace{-13pt}
\end{figure*}

\begin{figure} [h]
		\centering
	\includegraphics[width=0.8\linewidth,height=0.8\linewidth]{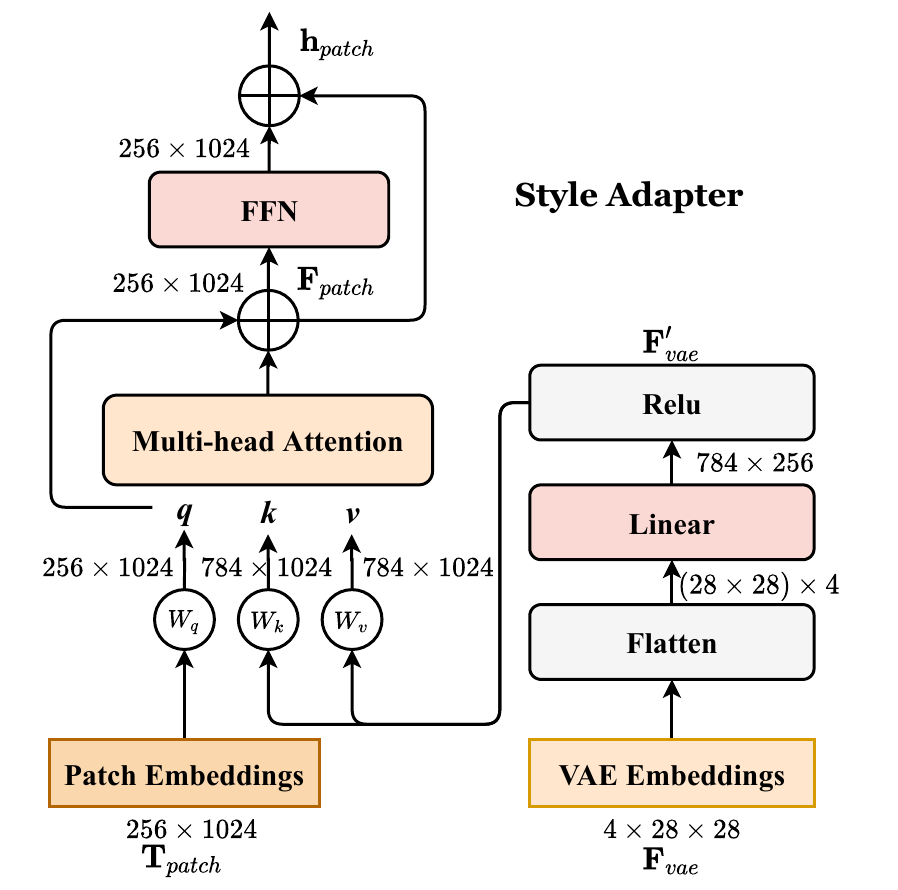}
		\caption{ The architecture of the style adapter. }
		\label{fig:SA}
  \vspace{-13pt}
	\end{figure}

In this paper, we propose a novel \textbf{TryOn-Adapter} to preserve the identity of the given garment while requiring relatively minimal training resources. The non-rigid warping preservation in virtual try-on is a challenging task. Previous diffusion-based methods~\cite{morelli2023ladi, gou2023taming, baldrati2023multimodal, kim2023stableviton, li2023warpdiffusion} do not decompose clothing identity adequately, resulting in unsatisfactory results, so we tackle this problem by dividing it into three factors, \ie, style, texture, and structure, and each factor is equipped with a special lightweight design: 
The Style Preserving module (Sec. \ref{Style Module}) aims to preserve the overall style of the garment. The Texture Highlighting module (Sec. \ref{THM and SAM}) focuses on refining high-frequency details. 
The Structure Adapting module (Sec. \ref{THM and SAM}) compensates for unnatural areas caused by clothing changes.
The T-RePaint (Sec. \ref{Repaint sec}) further reinforces the clothing identity preserving without compromising the overall image fidelity during the inference.

Specifically, as illustrated in Fig.~\ref{fig:main}, when given an image ${I_p}\in{\mathbb{R} ^{3\times H\times W}}$ of a person and an image $I_c\in{\mathbb{R} ^{3\times H'\times W'}}$ of a target garment, our method aims to generate an image $\hat{I}\in{\mathbb{R} ^{3\times H\times W}}$, where the person in ${I_p}$ is depicted wearing the garment from $I_c$. 
For input preprocessing, we first remove the original clothing from $I_p$ using a provided garment mask $m\in{\{0,1\} ^{1\times H\times W}}$, resulting in a clothing-agnostic RGB image $I_a\in{\mathbb{R} ^{3\times H\times W}}$, which retains non-target regions such as head and background. To obtain the warped garment image $I_c^w\in{\mathbb{R} ^{3\times H'\times W'}}$ and its mask $I_m^w\in{\{0,1\} ^{1\times H'\times W'}}$, we employ the garment warping model which here we choose {GP-VTON~\cite{xie2023gp}} to warp the original garment image $I_c$ into a coarse try-on shape based on the person's pose and other mask information in $I_p$. 
For the denoising process, the UNet model takes the pixel-wise addition of coarse warped garment image $I_c^w$ and the clothing-agnostic image $I_a$, along with the noisy person $I_t$ and mask $m$, as inputs. Besides, the target garment $I_{c}$, high-frequency map $I_{HF}$ extract from $I_c^w$, and human segmentation map $I_{seg}$ respectively serve as representations of style, texture, and structure, enabling fine-grained identity control. 
Note that our TryOn-Adapter is trained under a self-reconstruction manner, where $I_c$ is the exact garment worn by $I_{p}$, and $I_{seg}$ is also obtained from $I_{p}$. During the inference, $I_c$ and the clothing on $I_p$ are different, and $I_{seg}$ is generated by the proposed precise yet user-friendly method. For the latent space reconstruction process, the independent Enhanced Latent Blending Module is inserted into the autoencoder to further maintain consistent visual quality (Sec. \ref{ELBM sec}).

\subsection{Style Preserving Module}
\label{Style Module}
For this module, we extract as much style information as possible from the reference image to inject into the denoising U-Net to control the overall style of the generated garment, including color and category information. 
First, we input garment image $I_c$ through the frozen CLIP visual encoder to get the class token feature $\textbf{T}_{cls} \in {\mathbb{R} ^{1\times 1024}}$ and patch tokens features $\textbf{T}_{patch} \in {\mathbb{R} ^{256\times 1024}}$. The former is used as a coarse condition 
and added several additional fully connected layers to decode this feature given by:
   \begin{equation}\label{}   \ \ \ \ \ \ \ \ \ \ \ \ \ \ \qquad \qquad
   \textbf{h}_{cls} = \mathrm{MLPs}(\textbf{T}_{cls}),
   \end{equation}
where $\textbf{h}_{cls} \in {\mathbb{R} ^{1\times 1024}}$. Besides, unlike common image editing tasks, virtual try-on needs to guarantee the identity of the reference image fully, so we further introduce the latter to supplement fined style cues. However, although these two features embody sufficient style cues, they are not sensitive to color information.  
To enhance the color perception of patch tokens and guide the alignment of CLIP features with the output domain of the diffusion model, we design a style adapter to fuse CLIP patch embeddings $\textbf{T}_{patch}$ and VAE visual embeddings $\textbf{F}_{vae} \in {\mathbb{R} ^{4\times 28 \times 28}}$, as shown in Fig~\ref{fig:SA}. Formally: 
\begin{eqnarray}  \ \ \ \ \ \ \
    &\textbf{F}_{patch} = \mathrm{MHA}(\textbf{T}_{patch},\textbf{F}_{vae}',\textbf{F}_{vae}') +\textbf{T}_{patch}, \\
    &\textbf{h}_{patch} = \mathrm{FFN}(\textbf{F}_{patch}) + \textbf{F}_{patch},
\end{eqnarray}
where the $\textbf{F}_{vae}$ is obtained by the reference image $I_c$ through the pre-trained Stable Diffusion VAE encoder and $\textbf{F}_{vae}'\in {\mathbb{R} ^{784\times1024}}$ is obtained by $\textbf{F}_{vae}$ through a series of flatten and mapping operations. Moreover, $\mathrm{MHA}$ and $\mathrm{FFN}$ indicate multi-head attention and feed-forward network. To inject the coarse feature $\textbf{h}_{cls}$ and fined feature $\textbf{h}_{patch}$ into UNet, we do not merge them and replace the text tokens in the original stable diffusion model, as it was considered a naive solution that impedes the network from understanding the content in the reference image and the connection to the source image, as mentioned in Paint-by-Example~\cite{yang2023paint}. Therefore, followed by GLIGEN~\cite{li2023gligen}, $\textbf{h}_{cls}$ and $\textbf{h}_{patch}$ are fed into the diffusion process through cross-attention and gated self-attention, respectively. We denote $\textbf{v} = [v_1,\dots,v_M]$ as the visual feature tokens of an image. Therefore, the attention block of our denoising U-net consists of three attention layers, as shown in Fig.~\ref{fig:main}, which can be written as:
\begin{eqnarray} \qquad
    &\textbf{v} = \textbf{v} + \mathrm{SelfAttn}(\textbf{v}), \\
    &\textbf{v} = \textbf{v} +\beta \cdot \tanh \left( \gamma \right)  \cdot \mathrm{SelfAttn}([\textbf{v},\textbf{h}_{patch}]),\\
    &\textbf{v} = \textbf{v} + \mathrm{CrossAttn}(\textbf{v},\textbf{h}_{cls}),
\end{eqnarray}
where $\gamma$ is a learnable scalar initialized as 0, and $\beta $ is a constant to balance the importance of the adapter layer. Through gated self-attention, we have effectively introduced the guidance of style features to the generation process while avoiding the destruction of pre-trained weights.

\subsection{Texture Highlighting Module and Structure Adapting Module}
\label{THM and SAM}
After the Style Preserving module, the discriminative style features have been combined into the UNet. However, they struggle to preserve complex textures, such as patterns and logos, and exhibit obvious artifacts during the clothing transition when the original and target clothing are significantly difference, as well as in some cases involving challenging poses and body shapes. To address these issues, it is crucial to integrate explicit spatial conditions and maintain consistency between these guidance features and the generated image features to achieve perfect preservation. Therefore, inspired by T2I-Adapter~\cite{mou2023t2i}, we introduce two lightweight designs, the Texture Highlighting module and the Structure Adapting module. As shown in Fig.~\ref{fig:main}, they incorporate the high-frequency texture information for texture refinement and utilize the human body segmentation map for unnatural transition areas correction. Both conditions are encoded into multi-scale features and injected into the intermediate features of the denoising U-Net for precise control.

\begin{figure} [t!]
		\centering
	\includegraphics[width=\linewidth,height=0.7\linewidth]{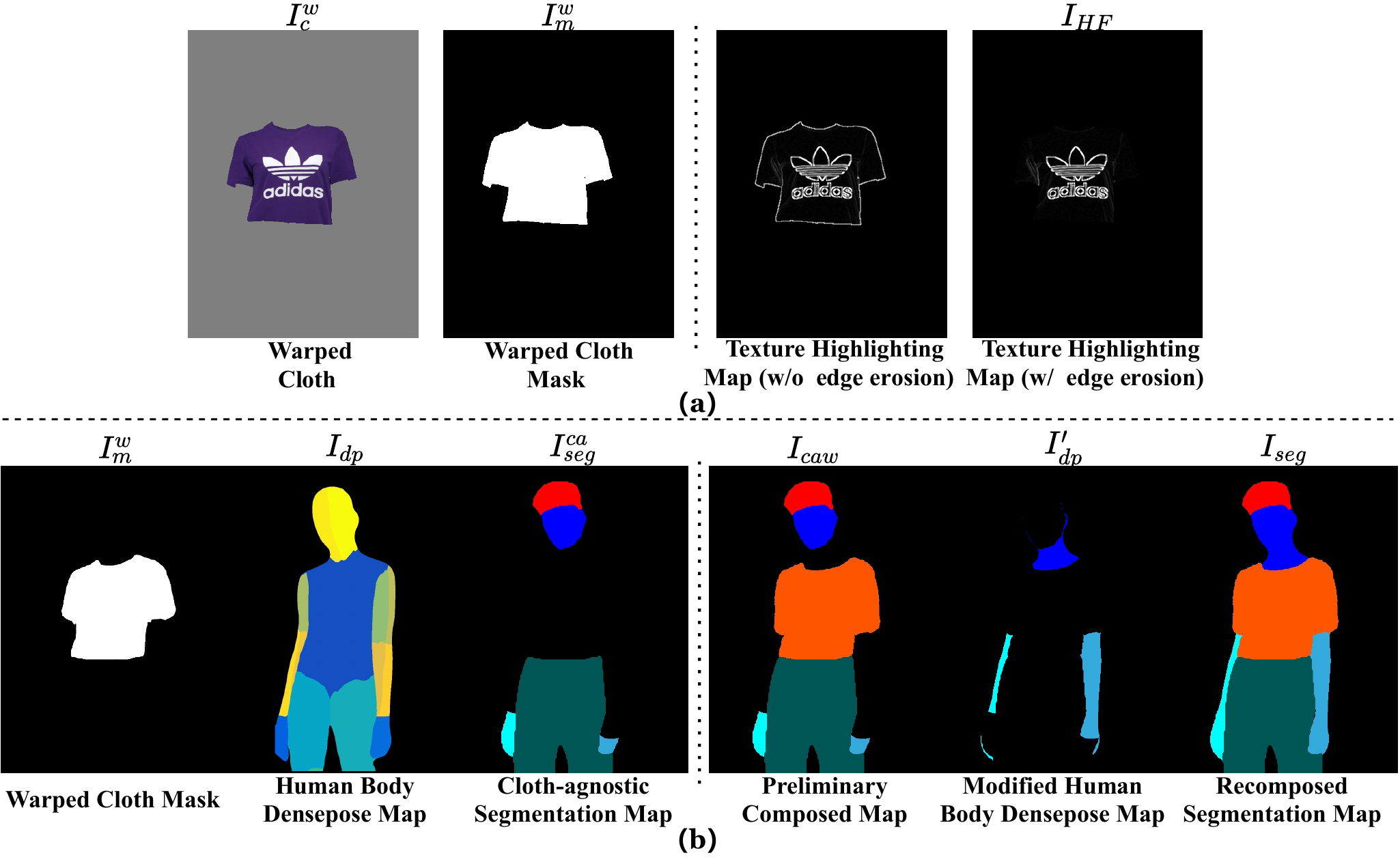}
   \vspace{-20pt}
		\caption{ (a): Visual illustration for the texture highlighting map generation. (b): Visual illustration for the target segmentation map generation.}
		\label{fig:illus}
  \vspace{-5pt}
	\end{figure}

For texture preservation, we extract the high-frequency map $I_{HF}$ of the warped garment by the sobel operator that highlights the complex texture and patterns of the garment, especially the logo and text. Besides, we observe that the edge information occasionally provides incorrect guidance since $I_{c}^{w}$ is just an offline rough result without adaptive refinement. To avoid introducing such ambiguous cues, we erode the edges of the clothes, given by:
 \begin{equation}\label{}  
I_{HF} = 0.5\times \left( \left| I_{c}^{w}\otimes \textbf{K}_{s}^{x} \right|+\left| I_{c}^{w}\otimes \textbf{K}_{s}^{y} \right| \right) \odot \left( I_{m}^{w}\ominus \textbf{K}_e \right),
   \end{equation}
where $\textbf{K}_{s}^{x}$, $\textbf{K}_{s}^{y}$, $\textbf{K}_e$ denote the horizontal, vertical Sobel kernels and erosion kernel. $\otimes$, $\odot$, $\ominus$ refer to convolution product, Hadamard product, and erosion operation. {The visual illustration for the texture highlighting map generation is as shown in Fig.~\ref{fig:illus}(a). }

For structure guidance, we utilize the segmentation map $I_{seg}$, which provides human posture information and explicitly indicates the clothing and body areas, serving as the strong prior information for correcting the discordant areas that appear after the garment change, such as the transition between long and short sleeves. { Unlike previous methods~\cite{choi2021viton,li2023virtual,xie2023gp, cui2023street} that use networks to predict the target segmentation map, this work avoids redundant off-the-shell networks, proposing a rule-based training-free segmentation extraction method to achieve precise results yet user-friendly process. The core idea of this design is to combine the existing cloth-agnostic segmentation map $I_{seg}^{ca}$, warped cloth Mask $I_{m}^{w}$, and the human body densepose~\cite{guler2018densepose} map $I_{dp}$ to obtain the decomposed segmentation map. Specifically, we first form a preliminary composed  image $I_{caw}$ by performing a per-pixel OR ($\lor 
$) operation to merge the $I_{seg}^{ca}$ with the $I_{m}^{w}$ in the binary logical space, \ie,  $I_{caw} = I_{seg}^{ca} \lor I_{m}^{w}$. Next, we combine $I_{caw}$ with the densepose map $I_{dp}$ to complete the missing arm parts. To remove the overlapping parts between $I_{caw}$ and $I_{dp}$ and the noise in $I_{dp}$ itself, we use the per-pixel AND ($\land$) operation and connectivity-based filtering ($\mathrm{Filter}_l$) to obtain the modified human pose map $I_{dp}'$.  This process removes noise and irrelevant details by excluding connected components with pixel counts below the threshold $l$ (here set to 12), given by:
 \begin{eqnarray}\label{}  \qquad \quad \
 &\mathrm{Filter}_l(\cdot )=\{p\in I|size(C_p)\ge l\}, \\
& I_{dp}^{\mathrm{'}}= \mathrm{Filter}_l(I_{dp}-(I_{dp}\land I_{caw})),
   \end{eqnarray}
where $I$ represents the image to be filtered, $p$ represents a pixel in the image, $C_p$ represents the connected region adjacent to pixel $p$, and $size(C_p)$ represents the number of pixels in the connected component.
Finally, the $I_{dp}'$ is merged with $I_{caw}$ through the per-pixel OR ($\lor$) operation to obtain the recomposed segmentation map $I_{seg}$, \ie, $I_{seg}=I_{caw} \lor I_{dp}' $. The visual illustration for the target segmentation map generation is shown in Fig.~\ref{fig:illus}(b).
}

\begin{figure} [t!]
		\centering
	\includegraphics[width=\linewidth,height=0.88\linewidth]{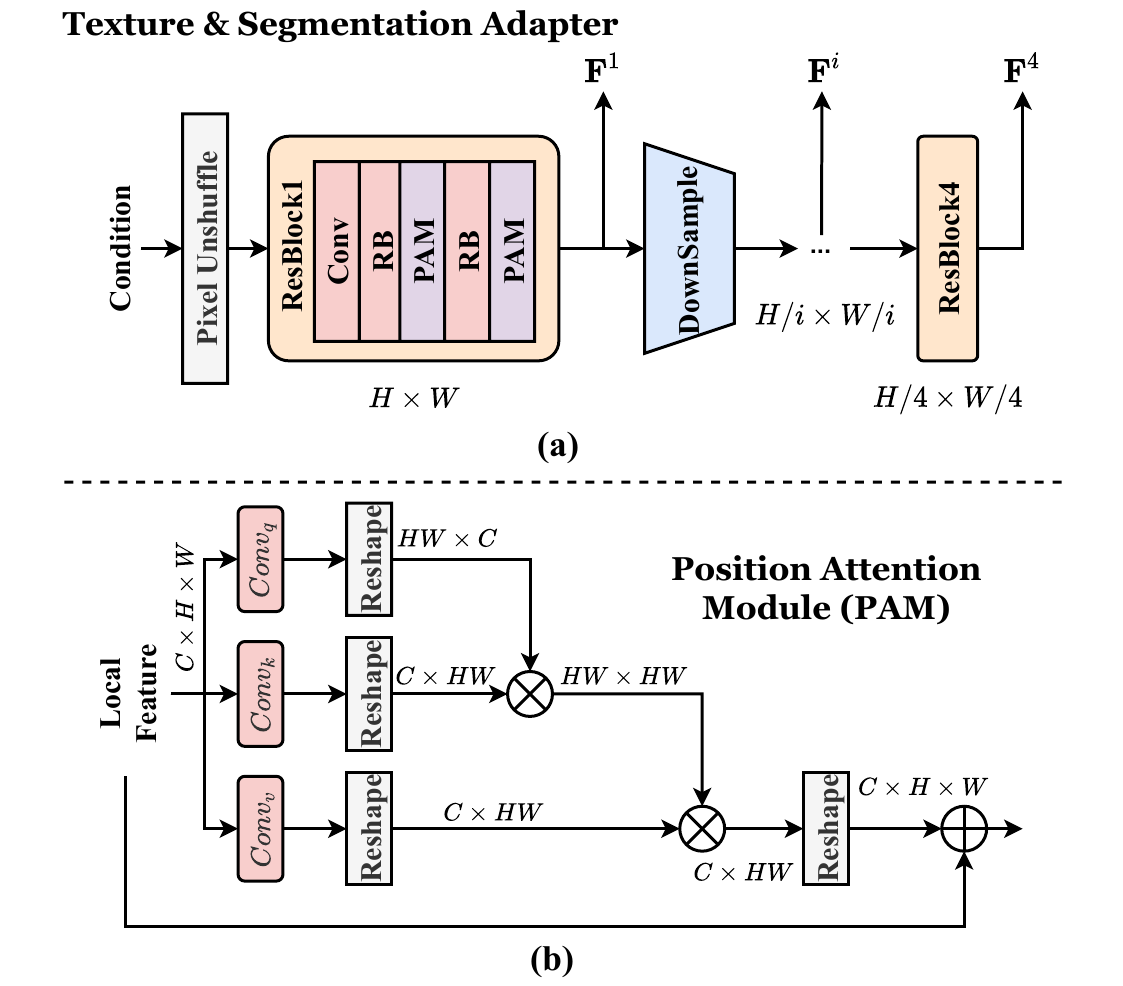}
    \vspace{-10pt}
		\caption{ (a): The architecture of the texture and segmentation adapter. Every ResBlock consists of a convolution layer, two resnet layers, and two position attention modules. (b): The architecture of the position attention module.}
		\label{fig:struc_ad}
  \vspace{-13pt}
	\end{figure}
In practice, we follow the network design in T2I-Adapter~\cite{mou2023t2i}, {and add Position Attention Modules (PAM) inspired by DANet~\cite{fu2019dual} to establish rich contextual relationships on local features in each resblock is shown in Fig~\ref{fig:struc_ad}(a). The architecture design of PAM is depicted in Fig.~\ref{fig:struc_ad}(b), which enhances the representation of spatial information for the texture highlighting map and the recomposed segmentation map.} Concretely, we introduce a Texture Adapter for the high-frequency map and a Segmentation Adapter for the segmentation map to obtain multi-scale conditional features $\textbf{F}_{HF}=\{\textbf{F}_{HF}^1,\textbf{F}_{HF}^2,\textbf{F}_{HF}^3,\textbf{F}_{HF}^4\}$, $\textbf{F}_{seg}=\{\textbf{F}_{seg}^1,\textbf{F}_{seg}^2,\textbf{F}_{seg}^3,\textbf{F}_{seg}^4\}$. These multi-scale features are correspond to the intermediate feature $\textbf{F}_{enc}=\{\textbf{F}_{enc}^1,\textbf{F}_{enc}^2,\textbf{F}_{enc}^3,\textbf{F}_{enc}^4\}$ in the denoising UNet encoder. Both adapters have the same network structure, as shown in Fig.~\ref{fig:struc_ad}(a). Finally, the conditional features $\textbf{F}_{HF}$, $\textbf{F}_{seg}$, and $\textbf{F}_{enc}$ are weighted and added at each scale to update $\textbf{F}_{enc}$, obtaining $\textbf{F}_{enc}'$ with:
 \begin{equation}\label{}  \qquad \quad \
 \textbf{F}_{enc}' = \textbf{F}_{enc} + \omega \cdot \textbf{F}_{seg} + (1-\omega) \cdot \textbf{F}_{HF},
   \end{equation}
where $ \omega \in (0,1)$ is a hyperparameter. The intermediate features of UNet are updated by injecting this explicit information, allowing it to focus on complex textural details and relationships of individual spatial parts.

\subsection{Diffusion Model for Virtual Try-On}
\label{Repaint sec}
In this work, we implement our method based on a pre-trained diffusion model built upon Stable Diffusion~\cite{rombach2022high}, \ie, Paint-by-Example~\cite{yang2023paint}, and added the identity preserving modules into this model to control the generation. The diffusion model includes two parts: an autoencoder (VAE), which can compress input images into latent space and reconstruct them, and a U-Net to perform denoising in the latent space directly. As shown in Fig.~\ref{fig:main}, for the first part, we embedded the ground-truth image $I_p$ and inpainting image $I_a'$ through the pre-trained encoder of VAE into the latent space, obtaining $\textbf{z}_0$ and $\textbf{z}_a'$. The forward process is executed at $\textbf{z}_0$ at a given timestamp $t$, with:
 \begin{equation}\label{forward process} \qquad \qquad \qquad \
\textbf{z}_t=\sqrt{\bar{\alpha}_t}\textbf{z}_0+\sqrt{1-\bar{\alpha}_t}\epsilon,
   \end{equation}
where $\textbf{z}_t$ indicates the noisy feature map at step $t$, $\alpha_t$ decreases with the timestep $t$, and $\epsilon \in \mathcal{N} \left( 0,\textbf{I} \right) 
$ is the Gaussian noise. For the generative process, we concatenate $\textbf{z}_t$,  $\textbf{z}_a'$, and the resized mask $m$ as the U-Net's input $\textbf{z}_t'=[\textbf{z}_t, \textbf{z}_a', m]$. The style features $\textbf{h}_c = [\textbf{h}_{cls},\textbf{h}_{patch}]$, texture condition $\textbf{F}_{HF}$, and structure guidance $\textbf{F}_{seg}$ are also injected into the UNet. Finally, our TryOn-Adapter is optimized via the objective:   
 \begin{equation}\label{}  
\mathbb{E} _{\textbf{z},t,\textbf{h}_c,\textbf{F}_{HF},\textbf{F}_{seg},\epsilon \in \mathcal{N} \left( 0,\textbf{I} \right)}\left[ \left\| \epsilon -\epsilon _{\theta }\left( \textbf{z}_t',{t},\textbf{h}_{c},\textbf{F}_{HF},\textbf{F}_{seg}\right) \right\| _{2}^{2}\right],
   \end{equation}
where the $\theta$ denotes the all learnable parameters.
\begin{figure} [t!]
		\centering
	\includegraphics[width=\linewidth,height=0.5\linewidth]{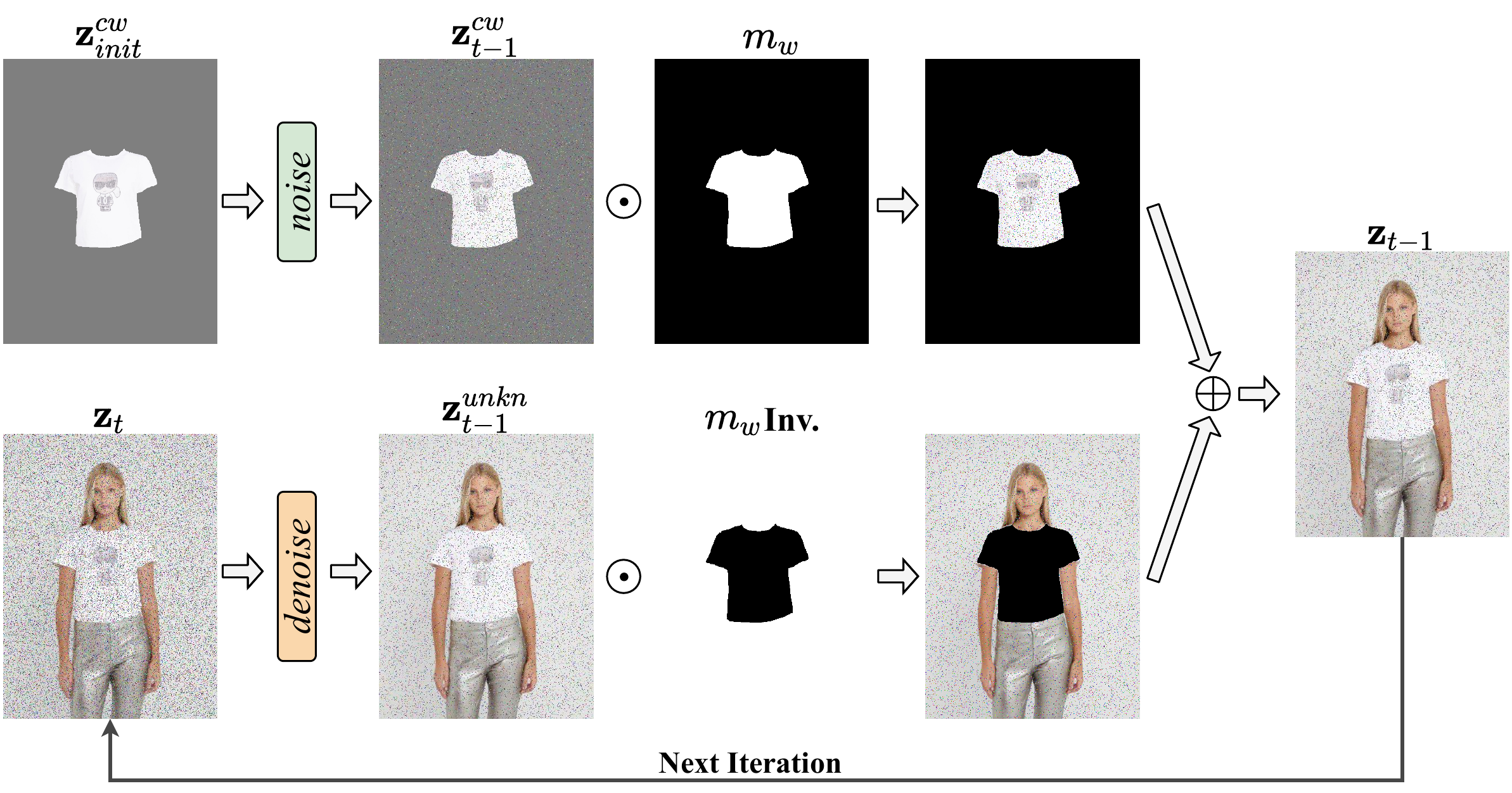}
		\caption{Overview of our T-RePaint for $ T'\leqslant t<T $.}
		\label{fig:repaint overview}
  \vspace{-13pt}
	\end{figure}
 
{To further reinforce the clothing identity preservation, inspired by previous works~\cite{lugmayr2022repaint,avrahami2023blended,corneanu2024latentpaint}, we utilize a training-free technique (\ie, RePaint) in the latent space during the inference.  RePaint is aimed at sampling known regions (\ie, unknown mappings) and replacing them at each denoising step in the inference process. Warped target garment images $I_c^w$ contain crucial prior information for preserving the identity, so applying RePaint to them further enhances the preservation effect. We observe that applying RePaint at all denoising steps results in noticeable noise at the RePaint edges and laacks realistic try-on effects in the final generated image. To address this problem, we propose a T-RePaint approach, applying RePaint only in the early denoising steps. Specifically, given a range of time steps $\left[ 1,T \right]$, the RePaint process starts from time step $T$ and ends on step $T' \ (T' < T)$. We feed $I_c^w$ to the VAE encoder to obtain the warped garment feature  $\textbf{z}^{cw}_{init}$, and the warped garment mask  $I_m^w$ is resized as $m_w$. We use $\textbf{z}_T$ to denote a noise sampled from the Gaussian distribution, and $\textbf{z}_0$ to denote the final image synthesis. Since the forward process is defined by Markov Chain at Eq.~\ref{forward process}, we can sample the warped garment feature at any time step $t$  to obtain the intermediate feature $\textbf{z}_{t-1}^{cw}$, \ie, $\mathbf{z}_{t-1}^{cw}\sim nosie\left( \mathbf{z}_{init}^{cw},t \right)$. Meanwhile, we use $\textbf{c}$ to denote all conditions in the denoising process based on the diffusion model, so the unknown regions' denoising at step $t$ can be defined as $\mathbf{z}_{t-1}^{unkn}\sim denosie\left( \textbf{z}_{t},\textbf{c},t \right)$. Thus, we achieve the reverse step with the composition of $\textbf{z}_{t-1}^{cw}$ and $\mathbf{z}_{t-1}^{unkn}$  controlled by the content keeping mask $m_w$, given by:
 \begin{equation}\label{}  
\left\{ \begin{array}{c}
	\textbf{z}_{t-1}=m_w\odot \textbf{z}_{t-1}^{cw}+\left( 1-m_w \right) \odot \mathbf{z}_{t-1}^{unkn}\,\,\left( T'\leqslant t<T \right)\\
	\textbf{z}_{t-1}=\textbf{z}_{t-1}^{unkn}\,\,\left( 1\leqslant t<T' \right)\\
\end{array}. \right. 
   \end{equation}
Our T-RePaint is shown in Fig.~\ref{fig:repaint overview} for $ T'\leqslant t<T $. }

\label{ELBM sec}
\begin{figure} [t!]
		\centering
	\includegraphics[width=\linewidth,height=0.65\linewidth]{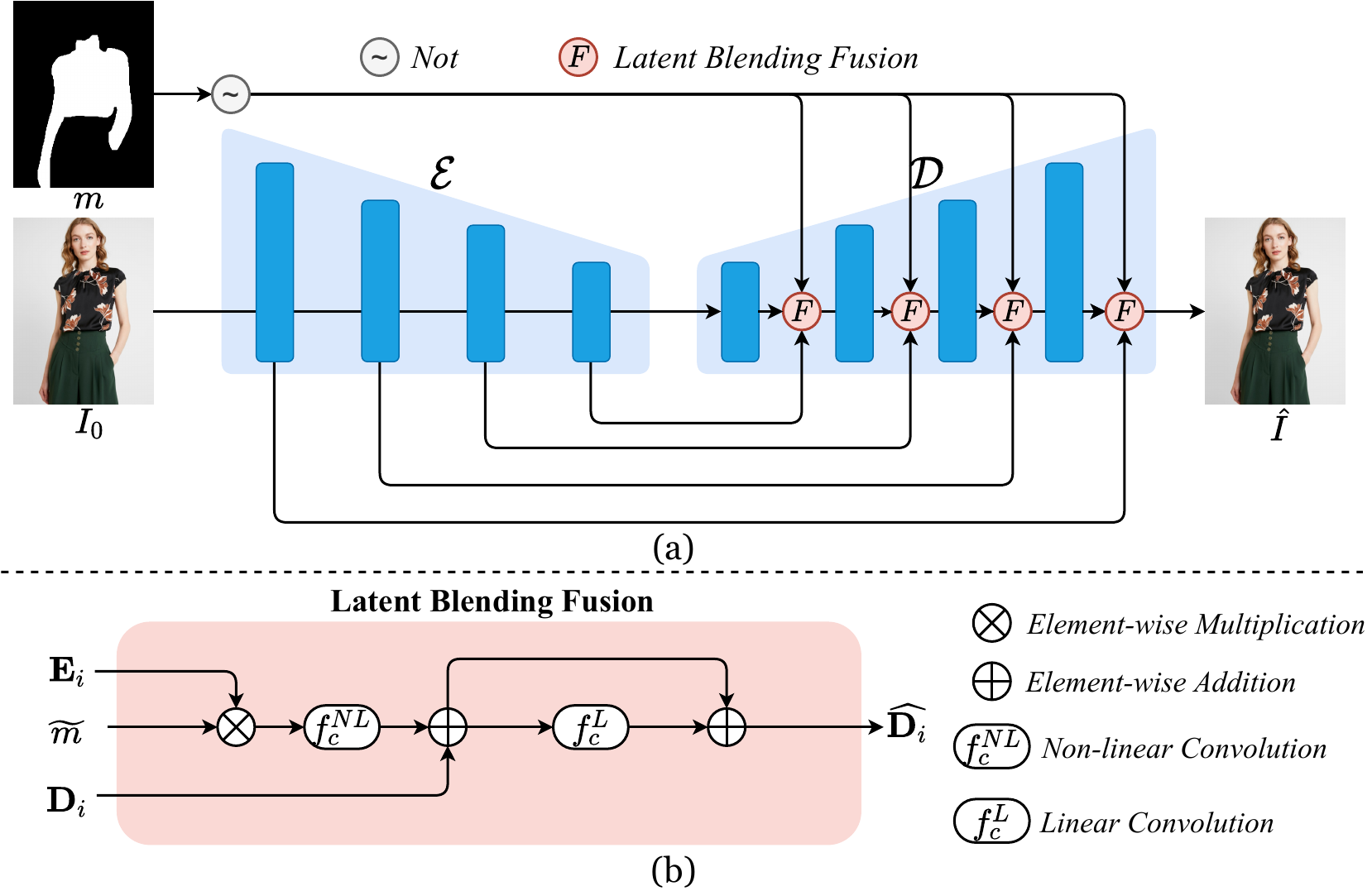}
		\caption{(a): Overview of our Enhanced Latent Blending Module. The autoencoder is frozen, and only the Latent Blending Fusion operation is learnable. (b): The architecture of Latent Blending Fusion operation.}
		\label{fig:ELBM}
  \vspace{-13pt}
	\end{figure}
 {
As mentioned before, the VAE enables the denoising network to operate in a lower-dimensional latent space, thereby reducing the computational cost in the diffusion network. However, due to data loss deriving from the spatial compression performed by the autoencoder, the latent space might struggle to capture high-frequency details precisely, which can easily lead to distortion of faces or hands in the generated images.
For the distortion problem, some previous methods~\cite{gou2023taming,li2023warpdiffusion} blend the background areas from the person image (\eg, face, hands) with the foreground areas (clothing) from the generated image at the pixel level, but bring about identifiable artifacts and blurred at the same time. By contrast, inspired by recent works~\cite{morelli2023ladi,li2019faceshifter,zhu2023designing, avrahami2023blended}, we propose the Enhanced Latend Blending Module (ELBM), which utilizes a background mask to directly copy the background region of the encoders' features from different layers and combines them with the corresponding ones of the decoder through some skip connections and learnable parameters. In this way, the VAE Decoder's difficulty in capturing high-frequency information is alleviated by blending enhanced background information into the decoding process. Specifically, we use $I_{0}$ to denote the original image and $m$ to denote the background mask. Given the encoder $\mathcal{E}$, the decoder $\mathcal{D}$ and the input $I_{0}$, the $i$-th feature map comes from the encoder and the decoder can be represented as $\textbf{E}_i$ and $\textbf{D}_i$, respectively. The enhanced latent blending process is formulated as:
 \begin{eqnarray}\label{EQ14}  \qquad \qquad \quad 
 &\widehat{\textbf{D}_i} =
 \textbf{D}_i + 
 f_{c}^{NL}\left( \textbf{E}_i\right)  \otimes \widetilde{m},  \\
 \label{EQ15}
  &\widehat{\textbf{D}_i} = \widehat{\textbf{D}_i} +  f_{c}^{L}\left( \widehat{\textbf{D}_i}\right),
   \end{eqnarray}
where $\otimes$ is element-wise multiplication, $\widetilde{m} = 1-m$. $f_{c}^{NL}$ and $f_{c}^{L}$ represent learnable non-linear and linear convolution.  {Unlike LaDI-VTON's~\cite{morelli2023ladi} EMASC, we further integrate the output $\widehat{\textbf{D}_i}$ of Eq.~\ref{EQ14} with a linear convolution and residual connection, as shown in Eq.~\ref{EQ15}, to reduce the probability of a disconnected feeling at the foreground-background junction.} The training process only employs a frozen autoencoder and trainable convolution layers under the supervision of reconstruction and VGG loss. Our ELBM is illustrated in Fig~\ref{fig:ELBM}. Through this design, the consistent visual quality of synthesized images has been significantly enhanced. 
}

%% file: sec/4_exp.tex
\section{Experiments}
\label{sec:exp}
\begin{figure*} [h!]
	\centering
	\includegraphics[width=0.95\linewidth]{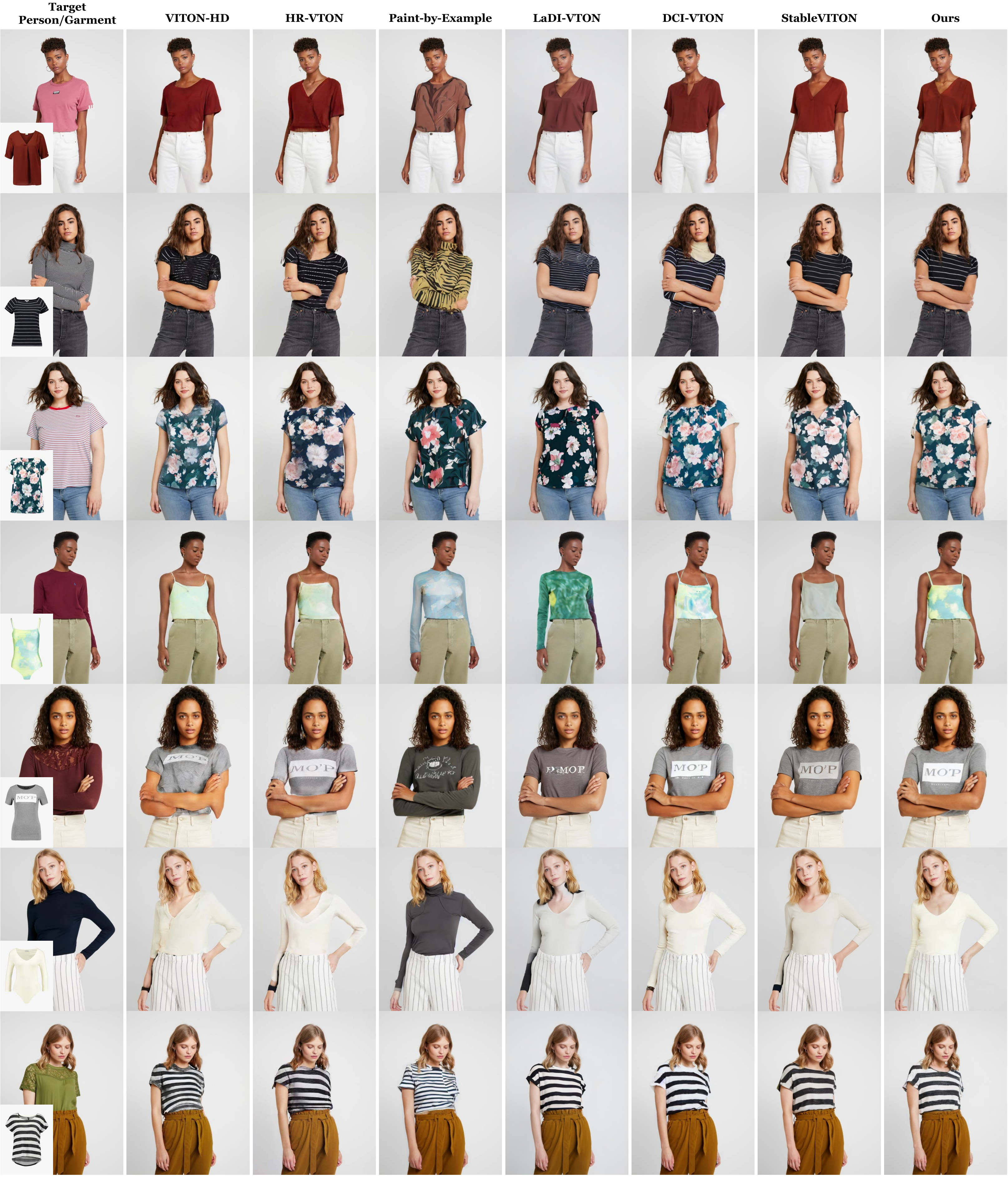}
	\caption{Qualitative comparison on the VITON-HD dataset~\cite{choi2021viton} with VITON-HD~\cite{choi2021viton}, HR-VTON~\cite{lee2022high}, Paint-by-Example~\cite{yang2023paint}, LaDI-VTON ~\cite{morelli2023ladi}, DCI-VTON~\cite{gou2023taming}, StableVITON~\cite{kim2023stableviton}, and our \textbf{TryOn-Adapter} at $512\times384$ resolution.}
	\label{fig:vis}
 \vspace{-9pt}
\end{figure*}

\begin{figure*} [h!]
	\centering
	\includegraphics[width=\linewidth]{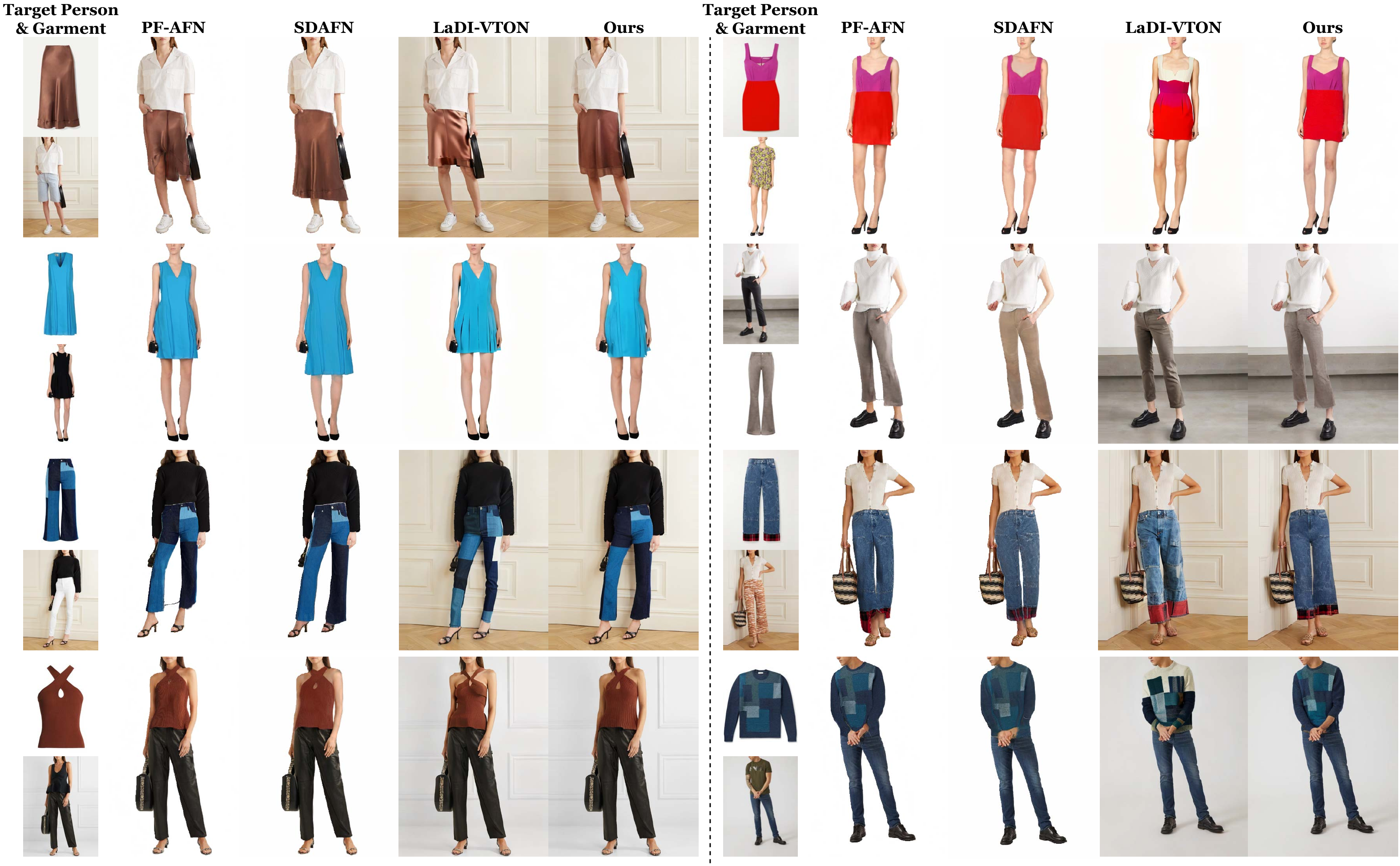}
	\caption{Qualitative comparison on the Dresscode dataset~\cite{morelli2022dress} with PF-AFN~\cite{ge2021parser}, SDAFN~\cite{bai2022single}, LaDI-VTON~\cite{morelli2023ladi}, and our \textbf{TryOn-Adapter} at $512\times384$ resolution.}
	\label{fig:dresscode}
 \vspace{-9pt}
\end{figure*}

\subsection{Experimental Setup}
\textbf{Datasets.} We mainly conduct qualitative and qualitative evaluations of our TryOn-Adapter on VITON-HD~\cite{han2018viton}, which comprises 13,679 image pairs. Each pair comprises a front-view upper-body woman and an upper garment under the resolution of $1024 \times 768$. Followed by the previous works~\cite{morelli2023ladi,gou2023taming,xie2023gp}, we split the dataset into 11,674/2,032 training/testing pairs. To prove that our method can have excellent results in more diverse scenarios, we further conduct experimental evaluations on Dreesscode~\cite{morelli2022dress}, which contains 53,792 front-view full-body person and garments pairs from different categories, \ie, upper, lower, and dresses. 


\noindent \textbf{Evaluation Metrics.} {To quantitatively evaluate our model, we use various metrics for the similarity and realism assessment. For similarity evaluation, we aim to assess the generated image's coherence compared to the ground truth, which can test the model's capability of ID preservation. This evaluation is mainly validated on paired images, for which we employed two widely used metrics: Structural Similarity (SSIM) for pixel level and Learned Perceptual Image Patch Similarity (LPIPS) for feature level. For realism assessment, the aim is to ensure that the generated images exhibit consistent visual quality and realistic try-on effects. Both paired images and unpaired images should be measured, and we use the Frechet Inception Distance (FID) and Kernel Inception Distance (KID) as our metrics at the feature level.}

\noindent \textbf{Implementation Details.}  We build our diffusion model based on Paint-by-Example~\cite{yang2023paint}, including an autoencoder with latent-space downsampling factor $f= 8$ and a UNet denoiser. We utilize its pre-trained model and freeze all parameters except attention layers. {We first train the ELBM module. For the diffusion model, the style preserving module is separately trained with the texture highlighting and structure adapting modules.}  We generate the images at $512 \times 384$ resolution, and the reference image $I_c$ is resized at $224 \times 224$.  We set $\omega=0.5$ in Sec.\ref{THM and SAM}. For optimizing, we utilize AdamW~\cite{loshchilov2017decoupled} optimizer with the learning rate of $1 \times 10^{-5}$, and we trained on 4 NVIDIA Tesla V100 GPUs for 40 epochs. For the inference, we utilize the PLMS~\cite{liu2022pseudo} sampling method, with 100 sampling steps, {and we set $T'=50$ in T-RePaint (see Sec.~\ref{Repaint sec})}.

\subsection{Quantitative and Qualitative Evaluations}
\begin{table*}[t!]
\centering
\caption{Quantitative comparisons on the VITON-HD dataset~\cite{choi2021viton}.The \textbf{boldfacen}  indicates the highest results. Note: * denotes results reported in previous works, which may differ in metric implementation. `Tunable Params' indicates the trainable parameters in the diffusion model. {Lacking open-source code to WarpDiffusion~\cite{li2023warpdiffusion}, the precise number of trainable parameters eludes us, yet their paper's full fine-tuning method implies it exceeds 859M.}}
    \setlength{\tabcolsep}{.4em}
 
\vspace{-5pt}
\begin{tabular}{cccccccccc} \hline & 
& {\textbf{Tunable}}& & & & & & \\ 
\multirow{-2}{*}{$\textbf{Method}$} & \multirow{-2}{*}{$\textbf{Reference}$} & \textbf{Params} &\multirow{-2}{*}{$\textbf{LPIPS}\downarrow$} &\multirow{-2}{*}{$\textbf{SSIM}\uparrow$}& \multirow{-2}{*}{$\textbf{FID}_\textbf{p}\downarrow$}&\multirow{-2}{*}{$\textbf{KID}_\textbf{p}\downarrow$} &\multirow{-2}{*}{$\textbf{FID}_\textbf{u}\downarrow$} &\multirow{-2}{*}{$\textbf{KID}_\textbf{u}\downarrow$} \\ \hline
VITON-HD~\cite{choi2021viton} & CVPR(21)& - & 0.116 & 0.863 & 11.01 & 3.71 & 12.96 & 4.09 \\
PF-AFN*~\cite{ge2021parser} &CVPR(21)& - & 0.087 & 0.886 & - & - & 9.48 &- \\
FS-VTON*~\cite{he2022style} &CVPR(22)& - & 0.091 & 0.883 & - & - & 9.55 &- \\
HR-VTON~\cite{lee2022high} & ECCV(22)& - & 0.097 & 0.878 & 10.88 & 4.48 & 13.06 & 4.72 \\ 
SDAFN*~\cite{bai2022single} &ECCV(22)& -  &0.092 & 0.882 & - & - & 9.40   &- \\
GP-VTON*~\cite{xie2023gp} & CVPR(23)& -  & 0.080 & 0.894 & - & - & 9.20 &- \\ \hline
TryOnDiffusion*~\cite{zhu2023tryondiffusion}& CVPR(23)&-  & - & - & - & - & 23.35 & 10.84 \\
Paint-by-Example~\cite{yang2023paint}&  CVPR(23)& 923M  & 0.143 & 0.843 & 9.97 & 1.72 & 11.04 & 2.09 \\
MGD*~\cite{baldrati2023multimodal} & ICCV(23) & 859M &  -&- & 10.60 & 3.26 &12.81 &3.86 \\
LaDI-VTON ~\cite{morelli2023ladi}&  ACMMM(23)& 1003M& 0.104 & 0.872 & 8.96 & 1.67& 9.93 & 1.91 \\
DCI-VTON~\cite{gou2023taming}& ACMMM(23)& 923M  & 0.072 & 0.892 & 5.57 & 0.57 & 8.76 & 0.87 \\ 
WarpDiffusion*~\cite{li2023warpdiffusion}& Arxiv(23)& $>$859M  & 0.078 & 0.896 & 8.90 & - & - & - \\ 
StableVITON~\cite{kim2023stableviton} & CVPR(24) & 611M  &0.082 & 0.865 &7.11&1.47 &9.76 &1.71 \\ 
StableVITON (RePaint)~\cite{kim2023stableviton} & CVPR(24) & 611M  &0.077 & 0.889 &6.17&1.06 &9.17 &1.32 \\ \hline
\textbf{TryOn-Adapter} & - &  \textbf{510M} & 0.071 & 0.894 & 5.57 & 0.56 & 8.63 & 0.79 \\
\textbf{TryOn-Adapter (RePaint)} & - &  \textbf{510M} &\textbf{0.069} & \textbf{0.897} & \textbf{5.54} & \textbf{0.53} & \textbf{8.62} & \textbf{0.78}  \\ \hline
\end{tabular}
\label{vton-hd}
\end{table*}

\begin{figure} [h!]
	\centering
	\includegraphics[width=\linewidth]{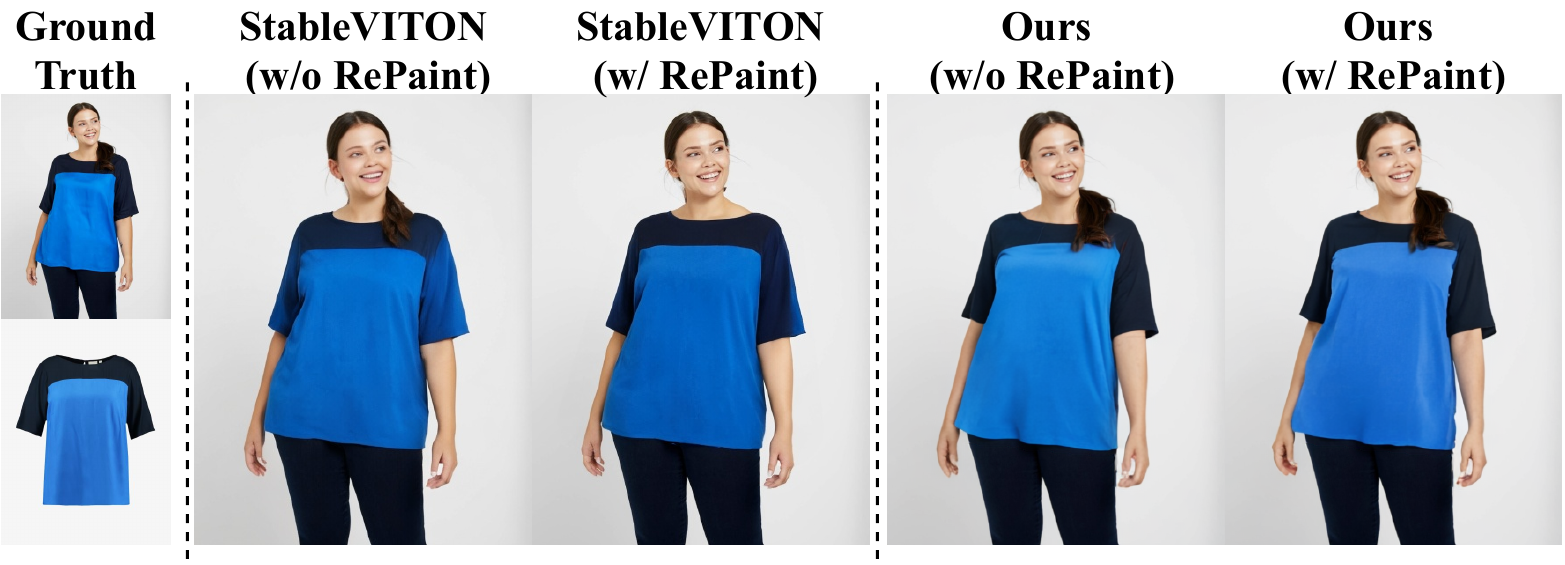}
 \vspace{-20pt}
	\caption{Qualitative evaluation on the VITON-HD dataset~\cite{choi2021viton} with StableVITON~\cite{kim2023stableviton} and our \textbf{TryOn-Adapter} at $512\times384$ resolution to compare the impact of RePaint on each method.}
	\label{fig:stableviton_vs}
 \vspace{-10pt}
\end{figure}

\begin{table*}[t]
    \caption{Quantitative results on the Dresscode dataset~\cite{morelli2022dress}.The \textbf{bold} indicates the highest results. Note: * denotes results reported in previous works, which may differ in metric implementation.\vspace{-.2cm}}
    \label{tab:dresscode}
    \setlength{\tabcolsep}{.4em}
    \resizebox{1.0\linewidth}{!}{
    \begin{tabular}{lc cc c cc c cc c cccccc}
    \hline
 & & \multicolumn{2}{c}{\textbf{Upper}} & & \multicolumn{2}{c}{\textbf{Lower}} & & \multicolumn{2}{c}{\textbf{Dresses}} & & \multicolumn{6}{c}{\textbf{All}} \\
    \cmidrule{3-4} \cmidrule{6-7} \cmidrule{9-10} \cmidrule{12-17}
    \textbf{Method} & & \textbf{FID$_\text{u}$} $\downarrow$ & \textbf{KID$_\text{u}$} $\downarrow$ & & \textbf{FID$_\text{u}$} $\downarrow$ & \textbf{KID$_\text{u}$} $\downarrow$ & & \textbf{FID$_\text{u}$} $\downarrow$ & \textbf{KID$_\text{u}$} $\downarrow$ & & \textbf{LPIPS} $\downarrow$ & \textbf{SSIM} $\uparrow$ & \textbf{FID$_\text{p}$} $\downarrow$ & \textbf{KID$_\text{p}$} $\downarrow$ & \textbf{FID$_\text{u}$} $\downarrow$ & \textbf{KID$_\text{u}$} $\downarrow$  \\ \hline
    PF-AFN*~\cite{ge2021parser} & & 14.32 & - & & 18.32 & - & & 13.59 & - & & - & - & - & - & - & - \\
    FS-VTON*~\cite{he2022style} & & 13.16 & - & & 17.99 & - & & 13.87 & - & & - & - & - & - & - & - \\
    HR-VTON*~\cite{lee2022high} & & 16.86 & - & & 22.81 & - & & 16.12 & - & & 0.086 & 0.901 & - & - & - & -  \\
    CP-VTON~\cite{wang2018toward} & & 48.31 & 35.25 & & 51.29 & 38.48 & & 25.94 & 15.81 & & 0.186 & 0.842 & 28.44 & 21.96 & 31.19 & 25.17 \\
    PSAD~\cite{morelli2022dress} & & 17.51 & 7.15 & & 19.68 & 8.90 & & 17.07 & 6.66 & & 0.058 & 0.918 & 8.01 & 4.90 & 10.61 & 6.17 \\
    SDAFN*~\cite{bai2022single}& & 12.61 & - & & 16.05 & - & & 11.80 & - & & 0.063 & 0.916 & - & - & - & - \\
    GP-VTON*~\cite{xie2023gp}& & 11.98 & - & & 16.07 & - & & 12.26 & - & & 0.050 & 0.925 & - & - & - & - \\ \hline
    LaDI-VTON ~\cite{morelli2023ladi}  & &{13.26} & {2.67} & & {14.80} & {3.13} & & {13.40} & {2.50} & & 0.064 & 0.906 & {4.14} & {1.21} & {6.48} & {2.20} \\
    MGD* ~\cite{baldrati2023multimodal} & &- & - & & - & - & & - & - & & - &- & 5.74 & 2.11 & 7.33 &2.82 \\
    WarpDiffusion*~\cite{li2023warpdiffusion} & &- & - & & - & - & & - & - & & 0.088 &0.895 & - & - & 8.61 &- \\ \hline
    \textbf{TryOn-Adapter} & &\textbf{11.58} & \textbf{1.63} & & \textbf{14.10} & \textbf{3.09} & & \textbf{11.58} & \textbf{1.66} & & \textbf{0.049} & \textbf{0.926} & \textbf{3.48} & \textbf{0.93} & \textbf{6.15} & \textbf{1.17} \\ \hline
    \end{tabular}
    }
\vspace{-.2cm}
\end{table*}

\begin{table*}[t!]
\centering
\caption{Effectiveness of our Adapter components on the VITON-HD dataset ~\cite{choi2021viton} at $512\times384$ resolution.  ``Params" and ``Tunable Params" indicate the total and trainable parameters in the diffusion model, respectively.}
\vspace{-5pt}
\begin{tabular}{ccccccc} \hline & 
& {\textbf{Tunable}}& & & &  \\

\multirow{-2}{*}{$\textbf{Method}$} & \multirow{-2}{*}{$\textbf{Params}$} & \textbf{Params} &\multirow{-2}{*}{$\textbf{LPIPS}\downarrow$} &\multirow{-2}{*}{$\textbf{SSIM}\uparrow$} &\multirow{-2}{*}{$\textbf{FID}_\textbf{u}\downarrow$} &\multirow{-2}{*}{$\textbf{KID}_\textbf{u}\downarrow$} \\ \hline
frozen & 859M & 0M &  0.227& 0.791 &  23.48 &  14.67  \\
frozen + fine-tuned attention layers & 859M &  84M  & 0.119  & 0.849  & 11.00  & 2.29    \\ \hline
 + style adaptation \ \ \ \ \  & 1048M  & 273M  & 0.079 & 0.887 & 8.89  & 0.94    \\  
 + texture adaptation \ & 1129M  & 354M  & 0.074 &  0.892 & 8.73 &   0.82 \\  
\ \ \ \ \ \ \ \ \  + segmentation adaptation & 1212M  & 435M  & 0.071 &0.894 &  8.63 &0.79   \\  \hline
\end{tabular}
\vspace{-10pt}

\label{ada}

\end{table*}
\vspace{-5pt}
\noindent \textbf{Quantitative Evaluations.}  
As shown in Tab.~\ref{vton-hd}, we quantitatively compare our method with the previous traditional methods on the VITON-HD dataset~\cite{choi2021viton}, including VITON-HD~\cite{choi2021viton}, PF-AFN~\cite{ge2021parser}, FS-VTON~\cite{he2022style}, HR-VTON~\cite{lee2022high}, SDAFN~\cite{bai2022single}, GP-VTON~\cite{xie2023gp}, and diffusion-based methods including TryOnDiffusion ~\cite{zhu2023tryondiffusion}, Paint-by-Example~\cite{yang2023paint}, MGD~\cite{baldrati2023multimodal}, LaDI-VTON ~\cite{morelli2023ladi}, DCI-VTON~\cite{gou2023taming}, WarpDiffusion~\cite{li2023warpdiffusion}, StableVITON~\cite{kim2023stableviton}. In traditional methods, GP-VTON~\cite{xie2023gp} has achieved the best performance, showing excellent structural similarity (SSIM) results. However, its performance in authenticity is not as good as the diffusion-based methods. {In full-tuning diffusion-based methods, due to the specified adaptation-based architecture for fine-grained identity factors, our method not only reduces the trainable parameters to nearly half compared to other methods but also achieves state-of-the-art performance across all metrics.} Besides, our method has seen a significant performance improvement compared to our baseline Paint-by-Example~\cite{yang2023paint}, thanks to the three identity-preserving modules we designed. {Additionally, in the unpaired setting, which is closer to real-world application scenarios, our KID and FID scores show significant advantages compared to other outperforming methods, such as DCI-VTON~\cite{gou2023taming}. Compared with the method StableVITON~\cite{kim2023stableviton}, which also employs the RePaint technique and efficient training, our method exhibits more excellent performance compared to StableVITON~\cite{kim2023stableviton} (rows 15, 17). Besides, the table results show that StableVITON's ability to preserve identity heavily relies on RePaint (rows 14, 15) even though it has more trainable parameters than ours. This also demonstrates that a single image is insufficient to fully capture the complexity of clothing identity. Conversely, our TryOn-Adapter itself has a strong ability to preserve garment identity (row 16), thanks to our decoupling of the identity preservation problem. Since our T-RePaint can bring some performance improvement and incurs no additional cost, we incorporate it into our approach (row 17). We also conduct qualitative comparisons to confirm this phenomenon, as shown in Fig.~\ref{fig:stableviton_vs}.}

{To further quantitatively evaluate our TryOn-Adapter, we compare our method on the Dresscode dataset~\cite{morelli2022dress} with the previous traditional methods, including PF-AFN~\cite{ge2021parser}, FS-VTON~\cite{he2022style}, HR-VTON~\cite{lee2022high}, SDAFN~\cite{bai2022single}, CP-VTON~\cite{wang2018toward}, PSAD~\cite{morelli2022dress},  SDAFN~\cite{bai2022single}, GP-VTON~\cite{xie2023gp}, and diffusion-based methods including  MGD~\cite{baldrati2023multimodal}, LaDI-VTON ~\cite{morelli2023ladi}, and WarpDiffusion~\cite{li2023warpdiffusion}. As shown in Tab.~\ref{tab:dresscode}, our TryOn-Adapter's performance has reached the most excellent results among all metrics under various settings.}  

\noindent \textbf{Qualitative Evaluations.} Fig.~\ref{fig:vis} shows the qualitative comparison of the results produced by different methods in the unpaired setting on the VITON-HD dataset~\cite{choi2021viton}. As depicted in the figure, although traditional methods like VITON-HD~\cite{choi2021viton} and HR-VTON~\cite{lee2022high} (as in columns 2 and 3) can preserve the identity of the target garment, the resulting garments exhibit some distortion when worn on a person, appearing unnatural. As for diffusion-based methods, the target garment can be worn naturally on a person, but it cannot guarantee the identity of the clothing. Paint-by-Example~\cite{yang2023paint} (as in column 4) and LaDI-VTON~\cite{morelli2023ladi} (as in column 5) cannot guarantee the style of the target garment, especially the color information. DCI-VTON~\cite{gou2023taming} compared to the previous two, has made great progress in style-preserving but has not effectively addressed the problem of long and short sleeves (as in column 6, rows 2 and 6), and the patterns and textures of the garments are not clear enough (as in column 6, rows 3 and 4). {Meanwhile,  StableVITON~\cite{kim2023stableviton} follows an efficient training strategy with ControlNet~\cite{zhang2023adding}, but it does not decouple the clothing identity preservation issue. This results in noticeable color discrepancies (as in column 7, rows 4, 5, and 6) and a lack of fidelity in fine texture details  (as in column 7, rows 4 and 5) compared to the target clothing in its output.} Compared to the above diffusion-based methods, our method benefits from the well-designed three adapter modules, effectively addressing the shortcomings. Consequently, our method can ensure a commendable preservation of garment identity (as in column 8, rows 1, 2, 4, and 7), featuring enhanced color fidelity (as in column 8), sharper illustration of intricate textures (as in column 8, rows 2, 3, 4, and 5), and better management of long/short sleeve transformations while naturally worn (as in column 8, rows 2, 4, and 6). 

{For further qualitative evaluations, we report in Fig.~\ref{fig:dresscode} sample images generated by our model and by the competitors using officially released weights on the Dresscode dataset~\cite{morelli2022dress}. Compared to traditional methods such as PF-AFN~\cite{ge2021parser} and SDAFN~\cite{bai2022single}, our method's try-on results will have a more realistic try-on effect without the unnatural signs of pasting from the warped garment onto the target person.  Compared to the diffusion-based method LaDI-VTON~\cite{morelli2023ladi}, our method has a distinct advantage in preserving the garment's identity, including elements like the style and texture details of the clothing.
}

\begin{figure*} [t!]
	\centering
 	\includegraphics[width=0.9\linewidth]{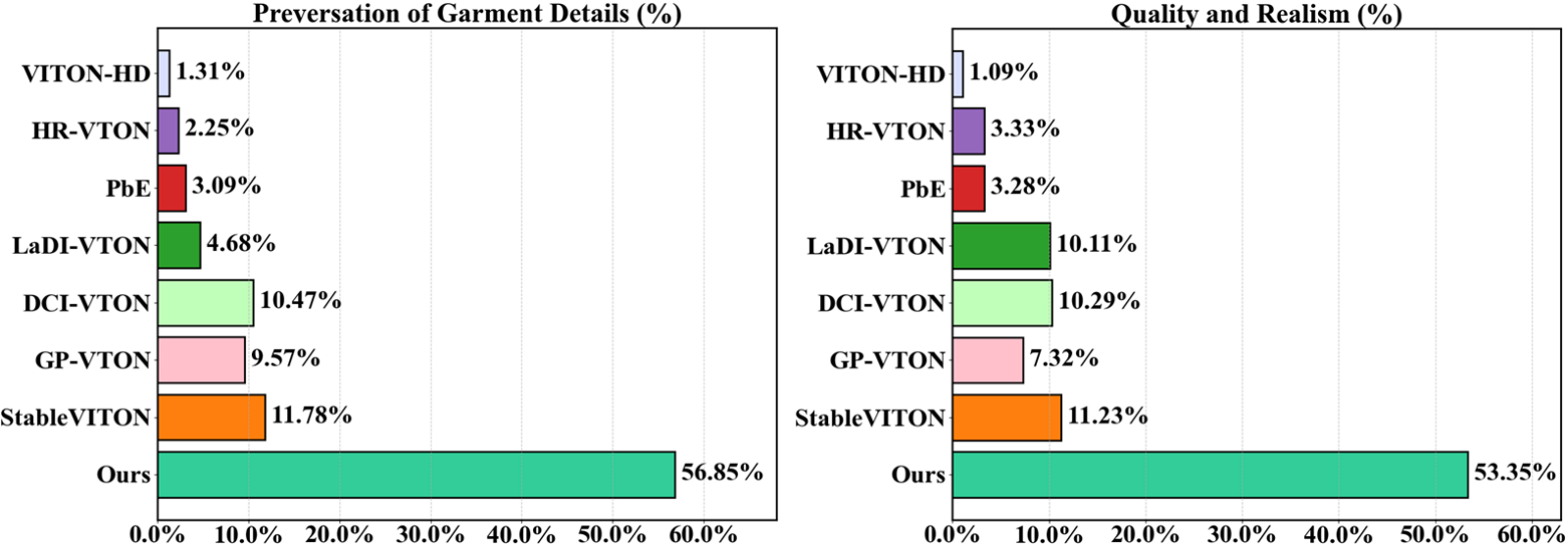}

 	\caption{User study results on VITON-HD dataset at $512 \times 384$ resolution. We compare our method with VITON-HD~\cite{choi2021viton}, HR-VTON~\cite{lee2022high}, Paint-by-Example (PbE)~\cite{yang2023paint}, LaDI-VTON ~\cite{morelli2023ladi},  DCI-VTON~\cite{gou2023taming}, GP-VTON~\cite{xie2023gp}, and StableVITON~\cite{kim2023stableviton}.}
	\label{fig:user}
\vspace{-15pt}
\end{figure*}

\noindent \textbf{User study of Virtual Try-On.} {We further evaluate our method against different methods, including  VITON-HD~\cite{choi2021viton}, HR-VTON~\cite{lee2022high}, Paint-by-Example (PbE)~\cite{yang2023paint}, LaDI-VTON ~\cite{morelli2023ladi}, DCI-VTON~\cite{gou2023taming}, GP-VTON~\cite{xie2023gp}, and StableVITON~\cite{kim2023stableviton} through a user study on different virtual try-on generation results in the VITON-HD dataset. We randomly select 300 unpaired sets from the test dataset, each containing a target garment image and a target person image. We survey 28 non-experts for this study, asking them to choose an image with the most satisfactory performance among the generated results of our model and baselines according to the following two questions: 1) Which image is the most photo-realistic? 2) Which image preserves the details of the target clothing the most? As shown in Fig.~\ref{fig:user}, our approach received over 50\% support for both questions. The results demonstrate that our method can generate naturally realistic images while effectively preserving target garment details during the virtual try-on process.}
\vspace{-5pt}

\subsection{Ablation Study and Further Analysis}
\begin{figure*} [h!]
	\centering
	\includegraphics[width=\linewidth]{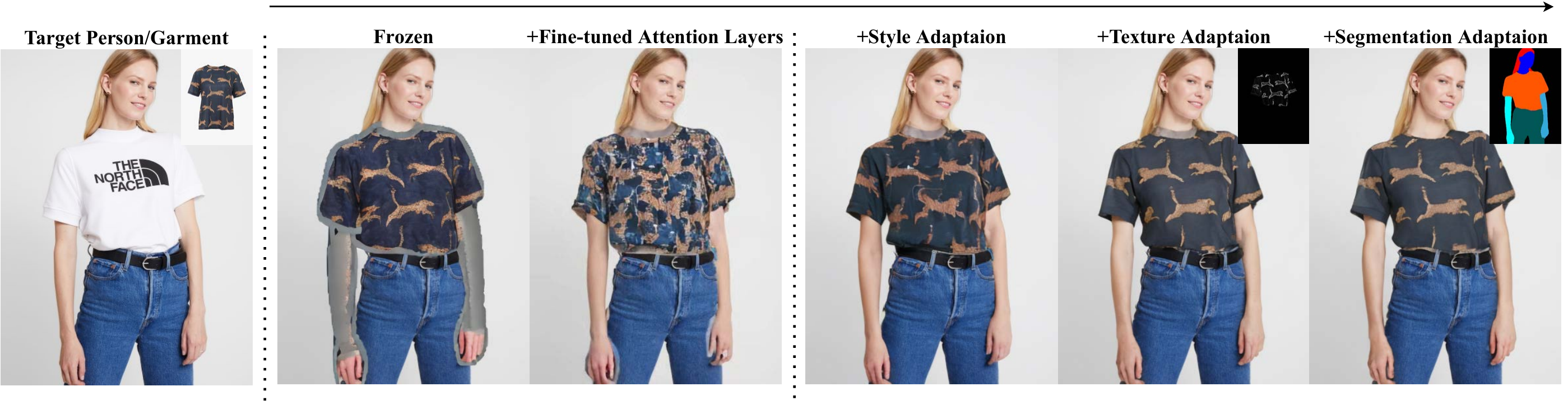}
 \vspace{-20pt}
	\caption{Visual effectiveness of individual adaptation components in our TryOn-Adapter on the VITON-HD dataset~\cite{choi2021viton}  at 512 $\times$ 384 resolution.}

	\label{fig:vis2}
   \vspace{-10pt}

\end{figure*}

\noindent \textbf{Effectiveness of Individual Adaptation Components.} To demonstrate the effectiveness of our proposed adaptation, we conduct ablation experiments on the VITON-HD~\cite{choi2021viton} dataset. {To more intuitively verify the effectiveness of each adaptation module, all results do not employ T-RePaint. Meanwhile, all tests use ELBM to prevent inaccuracies from reconstruction affecting result comparisons.
} We choose two baselines for comparison. One freezes all training parameters and uses Paint-by-Example's~\cite{yang2023paint} pre-trained model for inference, while the other is based on the former but only trains the attention layers and the CLIP class token's linear mapping layer related to the cross attention. As shown in Tab.~\ref{ada}, we gradually incorporate our designed adaptations, and the model's performance strengthens step by step. As shown in Fig.~\ref{fig:vis2}, the visual comparison of our 
generated results for each stage will be more intuitive. The frozen baseline is a semi-finished result (column 2), where the garment is detached from the body. For the second baseline (column 3), which is only fine-tuned on the attention layers, the generated clothing style diverges from the target clothing, and the boundary between the limbs and the garment is unclear. After adding the style adaptation (column 4), the clothing can naturally be worn on the person, and the clothing style has been significantly improved, but the details and textures of the clothing are not clear enough, and the shadow exists in the neck area. After combining texture adaptation (column 5), the representation of the clothing's detailed texture has been enhanced, but the high-frequency map lacks the ability to determine whether the shadow on the neck area is skin or a collar. After introducing segmentation adaptation (column 6), the neck shadow issue was successfully resolved.

\begin{table}[t!]
\centering
\caption{Quantitative analysis of the Style Adapter on the VITON-HD dataset~\cite{choi2021viton} at 512 $\times$ 384 resolution.} 
    \setlength{\tabcolsep}{.4em}

\vspace{-5pt}
{\begin{tabular}{ccccc}
\hline
    \textbf{Method} & {$\textbf{LPIPS}\downarrow$}   & {$\textbf{SSIM}\uparrow$} & {$\textbf{FID}_\textbf{u}\downarrow$}  & {$\textbf{KID}_\textbf{u}\downarrow$} \\ \hline
    w/o style adapter &   0.073  &  0.892  & 8.69 & 0.81  \\ 
    \textbf{Ours}	&\textbf{0.069} &  \textbf{0.897}   &  \textbf{8.62} & \textbf{0.78}  \\ \hline
\end{tabular}}
\label{style_tab}
\vspace{-10pt}
\end{table}

\begin{figure} [h!]
	\centering
	\includegraphics[width=\linewidth]{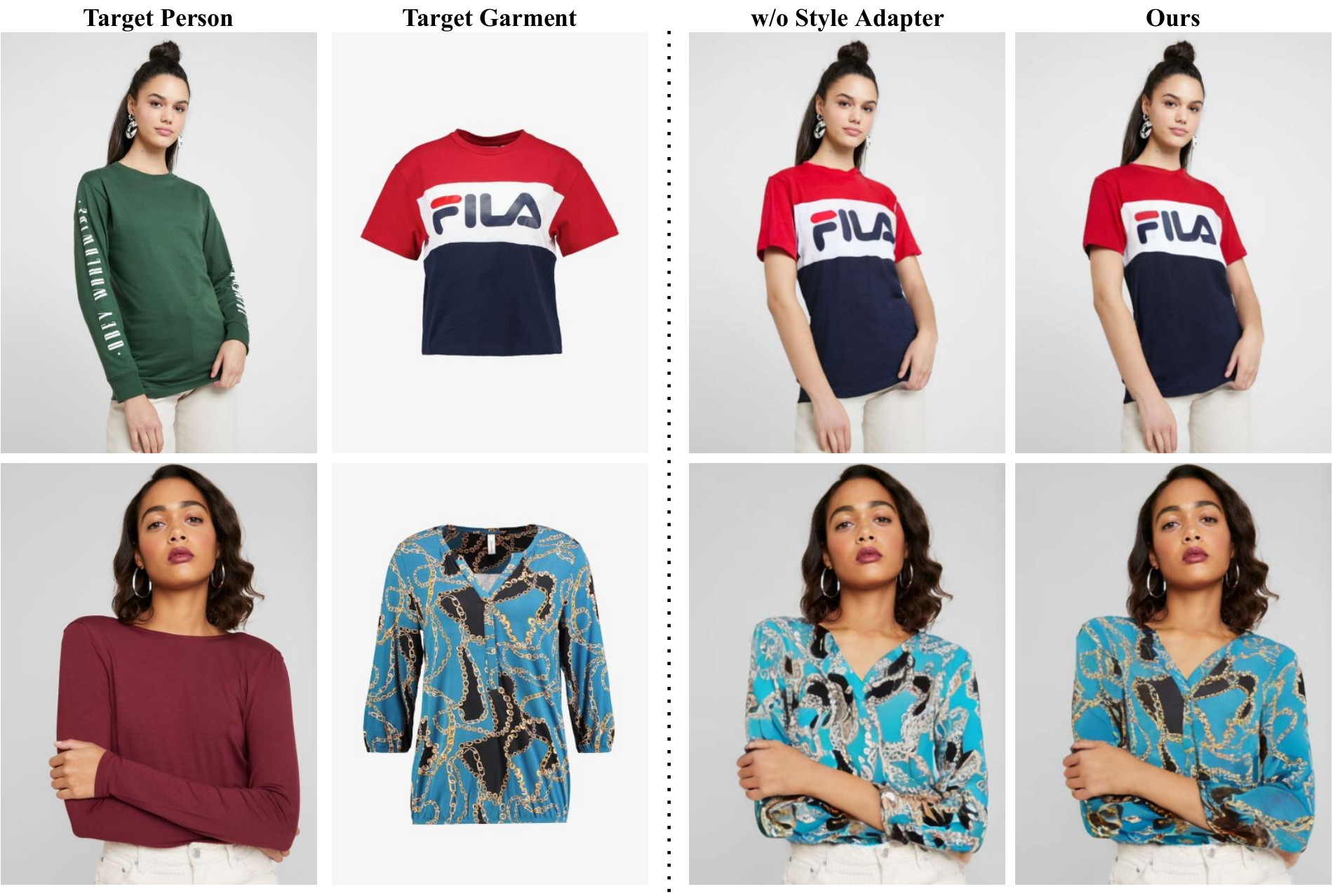}
	\caption{Visual effectiveness of the Style Adapter on the VITON-HD dataset~\cite{choi2021viton}  at 512 $\times$ 384 resolution.}
	\label{fig:style}
 \vspace{-15pt}
\end{figure}

\noindent \textbf{Analysis on Style Adapter.} To analyze the impact of our style adapter designed for patch token in the Style Preserving module, we conduct experimental evaluations on the VITON-HD dataset~\cite{choi2021viton}. As shown in Tab.~\ref{style_tab}, with the addition of the style adapter, all quantitative metrics have been improved.  {For a clear visual representation comparison on qualitative evaluation, we maintain consistency with previous ablation studies here by using ELBM and not T-RePaint.} As shown in Fig.~\ref{fig:style}, it can be seen that after adding the style adapter, there has been a noticeable improvement in the color difference between the generated garment (column 4) and the target garment (column 2) compared to the generated results without the style adapter (column 3). Meanwhile, the logos and textures on the garment also became clearer after integrating this module. The above results demonstrate the significance of integrating VAE embeddings, while also proving the effectiveness of our style adapter. 

\begin{figure*} [h!]
	\centering
	\includegraphics[width=\linewidth]{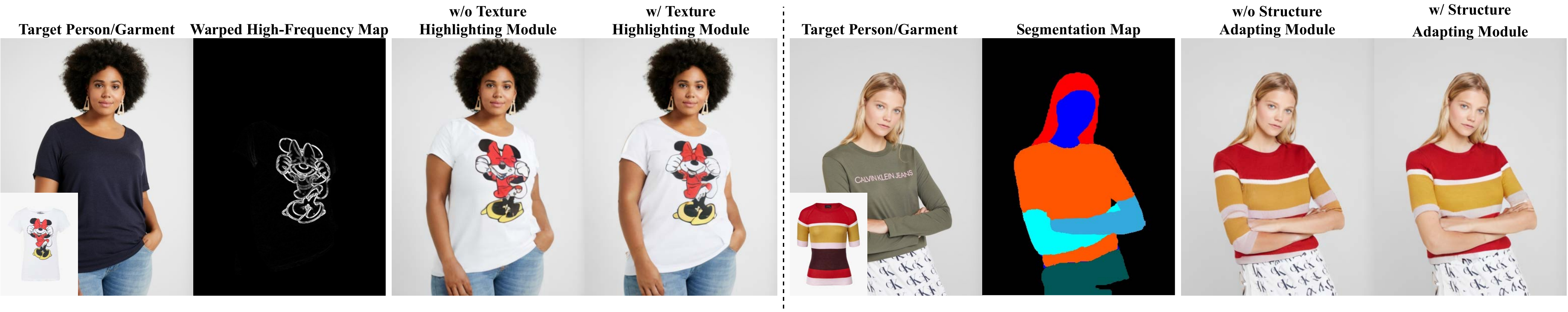}
 \vspace{-20pt}
	\caption{Qualitative evaluation of Texture Highlighting Module and Structure Adapting Module in our TryOn-Adapter on the VITON-HD dataset ~\cite{choi2021viton}  at 512 $\times$ 384 resolution.}

	\label{fig:module}
   \vspace{-10pt}

\end{figure*}
\noindent \textbf{Qualitative Evaluation of Texture Highlighting Module and Structure Adapting Module.} {To verify the robustness of Texture Highlighting Module and Structure Adapting Module, we supply more convincing visual results as shown in Fig.~\ref{fig:module}. Here, we also use ELBM and do not use T-RePaint. The example on the left in this figure proves that the Texture Highlighting Module can effectively enhance the texture of the target garment, especially the details of cartoon patterns. And, the example on the right demonstrates that the Structure Adapting Module is capable of addressing the problem of long and short sleeves well, bringing about a realistic try-on effect. 
}

\begin{figure*} [th!]
	\centering
	\includegraphics[width=\linewidth]{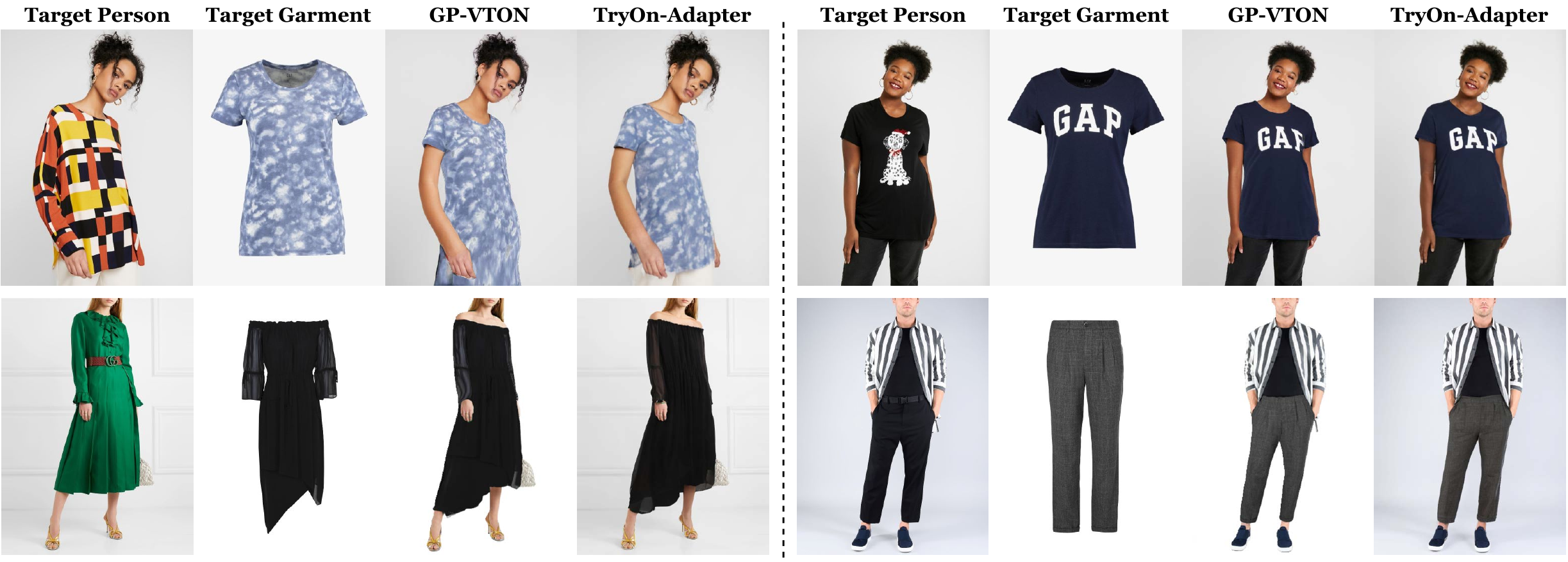}
 \vspace{-20pt}
	\caption{Qualitative Comparison between the Diffusion Model and GAN (GP-VTON~\cite{xie2023gp} $vs.$ TryOn-Adapter) on the VITON-HD~\cite{choi2021viton} and Dresscode ~\cite{morelli2022dress} datasets at 512 $\times$ 384 resolution.}

	\label{fig:gpvton}
   \vspace{-10pt}
\end{figure*}

\noindent \textbf{Qualitative Comparison between the Diffusion Model and GANs.} {To analyze the performance of the Diffusion Model and GAN in virtual try-on, we compare our TryOn-Adapter with the latest GANs-based method GP-VTON~\cite{xie2023gp}, where both use the same warped garment and the latter employs the GANs as the generative model as shown in Fig.~\ref{fig:gpvton}. The try-on results generated by GP-VTON~\cite{xie2023gp} (based on GANs) are prone to distortions and deformations, as seen in the waist area of the first row's left image and the logo area of the right image. Furthermore, the outputs from GP-VTON
lack a realistic sense of actual try-on, resembling a warped garment pasted onto the target person, as in the second row, especially with the arm in the sleeve on the left image, which is almost completely forgotten. Besides, GP-VTON exhibits noticeable jaggies around the clothing when zoomed in. Therefore, diffusion models possess more powerful generative capabilities than GANs.
}

\begin{table}[t!]
\centering
\caption{Quantitative analysis of PAM in Texture \& Segmentation Adapter on the VITON-HD dataset~\cite{choi2021viton} at 512 $\times$ 384 resolution.} 
    \setlength{\tabcolsep}{.4em}

\vspace{-5pt}
{\begin{tabular}{ccccc}
\hline
    \textbf{Method} & {$\textbf{LPIPS}\downarrow$}   & {$\textbf{SSIM}\uparrow$} & {$\textbf{FID}_\textbf{u}\downarrow$}  & {$\textbf{KID}_\textbf{u}\downarrow$} \\ \hline
    w/o PAM &   0.071  &  0.895  & 8.65 & 0.80  \\     \textbf{Ours}	&\textbf{0.069} &  \textbf{0.897}   &  \textbf{8.62} & \textbf{0.78}  \\ \hline
\end{tabular}}
\label{PAM_tab}
\vspace{-5pt}
\end{table}
\begin{figure} [h!]
	\centering
	\includegraphics[width=\linewidth]{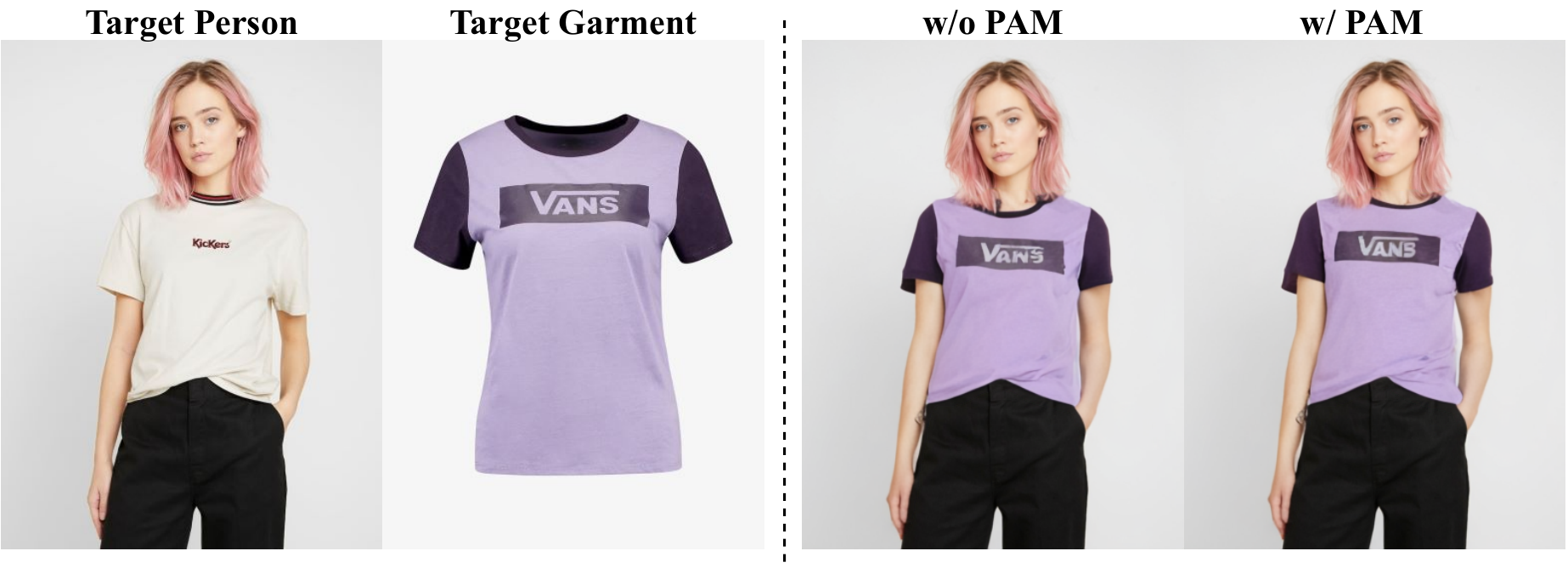}
  \vspace{-20pt}
	\caption{Quantitative evaluation of PAM in Texture \& Segmentation Adapter on the VITON-HD dataset~\cite{choi2021viton} at 512 $\times$ 384 resolution.}
	\label{fig:PAM}
 \vspace{-15pt}
\end{figure}
\noindent \textbf{Analysis on PAM in Texture \& Segmentation Adapter.} {To analyze the impact of the position attention module (PAM) in Texture \& Segmentation Adapter, we conduct experiments on the VITON-HD dataset~\cite{choi2021viton}. For the quantitative evaluation, we can see all quantitative metrics are improved after adding PAM, as shown in Tab.~\ref{PAM_tab}. For the qualitative evaluation, we don't use T-RePaint for a direct visual comparison. As shown in Fig.~\ref{fig:PAM}, the logos in the generated images are more evident after integrating PAM, benefiting from the enhanced local spatial representation by PAM, which allows the Adapters to interpret the high-frequency information in the images better.
}

\begin{figure} [t!]
	\centering
	\includegraphics[width=0.85\linewidth]{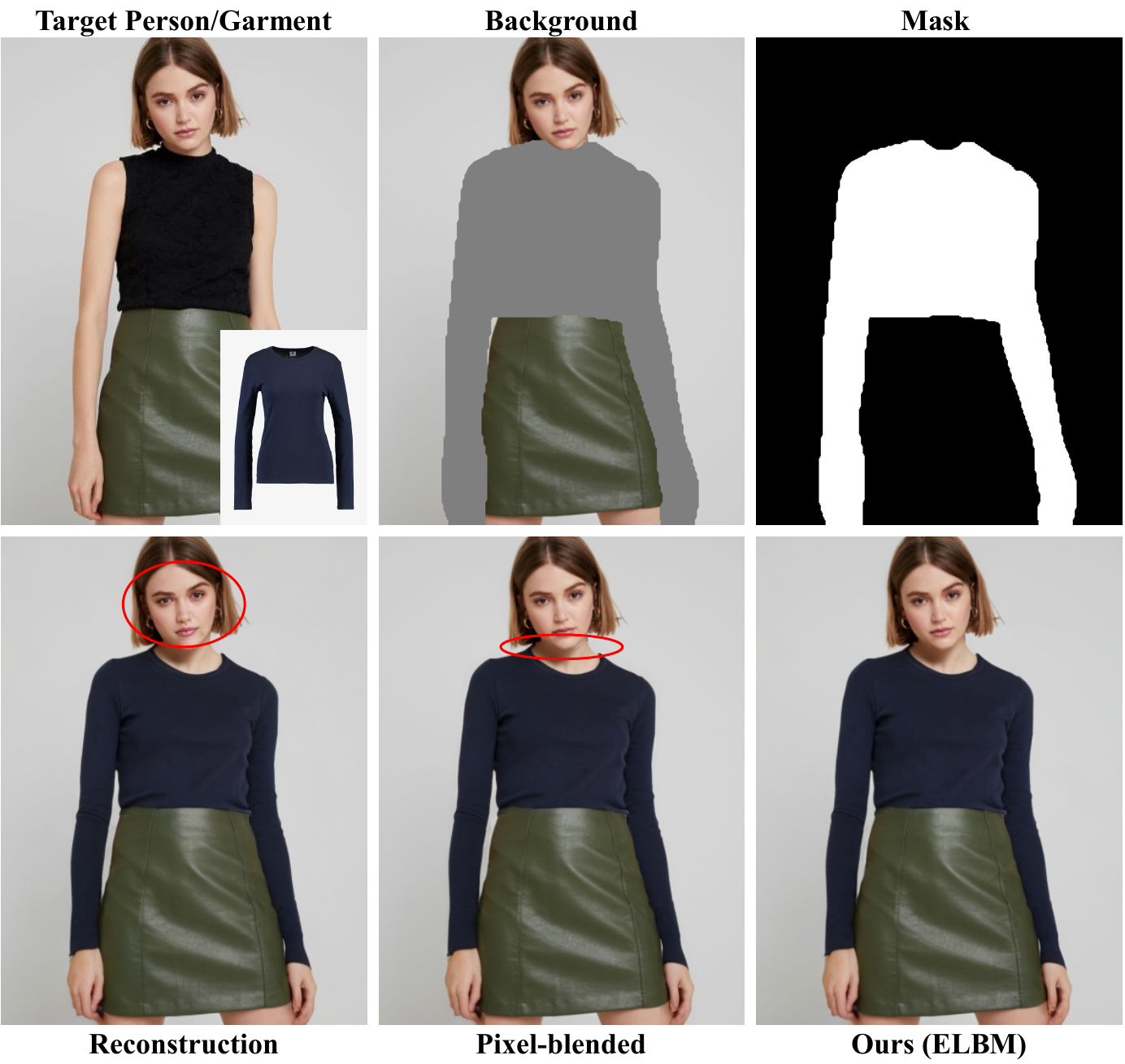}
  \vspace{-10pt}
	\caption{Qualitative evaluation of Enhanced Latent Blending Module (ELBM) on the VITON-HD dataset~\cite{choi2021viton} at 512 $\times$ 384 resolution. The reconstruction result is from the VAE of Stable Diffusion, and the pixel-blended result is from the combination of the original target person image and the reconstruction result image at the pixel level.}
	\label{fig:ELBM_abl}
 \vspace{-15pt}
\end{figure}

\begin{figure*} [h!]
	\centering
	\includegraphics[width=\linewidth]{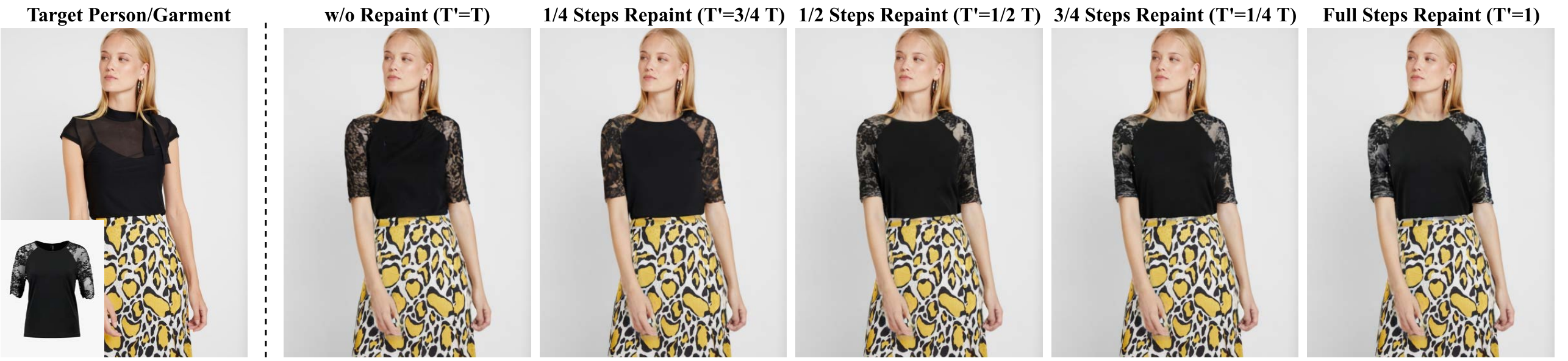}
  \vspace{-20pt}
	\caption{Qualitative evaluation comparing the impact of varying numbers of RePaint steps on the VITON-HD dataset~\cite{choi2021viton} at 512 $\times$ 384 resolution.}
	\label{fig:repaint}
 \vspace{-15pt}
\end{figure*}

\begin{table}[t!]
\centering
\caption{Quantitative analysis of PAM in Texture \& Segmentation Adapter on the VITON-HD dataset~\cite{choi2021viton} at 512 $\times$ 384 resolution.} 
    \setlength{\tabcolsep}{.4em}
\vspace{-5pt}
{\begin{tabular}{cccccc}
\hline
    \textbf{Task}   & \textbf{ELMB} & $f_c^{NL}$ & $f_c^{L}$&   {$\textbf{LPIPS}\downarrow$} & {$\textbf{SSIM}\uparrow$} \\ \hline
    Reconstruction&   w/o  &  $\usym{2717}$  & $\usym{2717}$ & 0.024 & 0.937  \\     
    Reconstruction  	&w/ &  $\usym{2713}$   &  $\usym{2717}$ & 0.021 & 0.954  \\ 
     Reconstruction	&w/ &  $\usym{2713}$   &  $\usym{2713}$ & \textbf{0.020} & \textbf{0.956} \\ \hline
       Try-On (paired)	&w/o &  $\usym{2717}$   &  $\usym{2717}$ & 0.076  & 0.867  \\
       Try-On  (paired)	&w/ &  $\usym{2713}$   &  $\usym{2717}$ & 0.071 & 0.895  \\ 
       Try-On (paired)	&w/ &  $\usym{2713}$   &  $\usym{2713}$ & \textbf{0.069} & \textbf{0.897} \\ \hline
     
\end{tabular}}
\label{ELBM_tab}
\vspace{-10pt}
\end{table}
\noindent \textbf{Analysis on Enhanced Latent Blending Module (ELBM).} 
{To analyze the impact of the Enhanced Latent Blending Module, we conduct qualitative and quantitative evaluations on the VITON-HD dataset~\cite{choi2021viton}. For the qualitative evaluation, we compare three approaches at the final image synthesis stage on virtual try-on: reconstruction, pixel-blended, and our ELBM. Given the original person image $I_0$ and background mask $m$, the reconstruction $I_{re}$ result is from the VAE of Stable Diffusion, and the pixel-blended result $I_{pb}$ is from the combination of the original target person image and the reconstruction result image at the pixel level, \ie, $I_{pb} = I_0 \otimes (1-m) + I_{re} \otimes m$. As shown in Fig.~\ref{fig:ELBM_abl}, the reconstruction result (column 1, row 2) exhibits some degree of distortion and deformation in the human face, whereas the pixel-blended result (column 2, row 2) preserves the critical facial features well but introduces noise and shadows at the junction of the neck due to the rough combination of $I_0$ and $I_{re}$. Our ELBM (column 3, row 2) effectively addresses the above problems, preserving high-frequency background information, such as the face and hands, while avoiding introducing any noise that may result from image combination. Please zoom in for more details.
}

{For the quantitative evaluation, we conduct experiments on two tasks, including image reconstruction and paired virtual try-on, and we ablate the impacts of our two convolutions $f_c^{NL}$ and  $f_c^{L}$ in latent blending fusion of ELBM.  As shown in Tab~\ref{ELBM_tab},  the ELBM we propose not only improves the reconstruction capabilities of the Stable Diffusion autoencoder in the reconstruction task but also elevates the overall performance of the final virtual try-on pipeline, resulting in superior evaluation metrics. {At the same time, the second and fifth rows in Tab.~\ref{ELBM_tab} represent LaDI-VTON~\cite{morelli2023ladi}, while the third and sixth rows represent our proposed ELBM. Our ELBM exhibits better performance, which demonstrates the effectiveness of deep fusion in Eq.~\ref{EQ15}.} 
}

\begin{table}[t!]
\centering
\caption{Quantitative evaluation comparing the impact of varying numbers of RePaint steps on the VITON-HD dataset~\cite{choi2021viton} at 512 $\times$ 384 resolution.} 
    \setlength{\tabcolsep}{.4em}

\vspace{-5pt}
\scalebox{1.2}{\begin{tabular}{ccccc}
\hline
    \textbf{Method} & {$\textbf{SSIM}\uparrow$}   & {$\textbf{LPIPS}\downarrow$} & {$\textbf{FID}_\textbf{u}\downarrow$}  & {$\textbf{KID}_\textbf{u}\downarrow$} \\ \hline
    w/o RePaint &   0.894  &  0.071  & 8.63 & 0.79  \\     
    1/4 Steps RePaint &0.896 &  0.070   &  \textbf{8.62} & \textbf{0.78}  \\ 
     1/2 Steps RePaint &\textbf{0.897} &  \textbf{0.069}   &  \textbf{8.62} & \textbf{0.78}  \\ 
      3/4 Steps RePaint & 0.895 & 0.071   &  8.67 & 0.82  \\
       Full Steps RePaint & 0.896 & 0.074   &  8.99 & 1.09  \\ \hline
    
\end{tabular}}
\label{repaint_tab}
\vspace{-10pt}
\end{table}

\noindent \textbf{Analysis on T-RePaint.} {To evaluate the impact of varying numbers of RePaint steps, we conduct quantitative and qualitative experiments on the VITON-HD dataset~\cite{choi2021viton}. For qualitative evaluation, utilizing RePaint for half of the denoising steps ($T' = 1/2~T$) during the inference achieves a balance between preserving the identity of the garment and realizing a realistic try-on effect, thereby attaining the best generative outcomes, as illustrated in Fig.~\ref{fig:repaint}.  Meanwhile, a larger T' yields a more realistic try-on effect but poorer texture ID preservation (see columns 2 and 3), and vice versa. Especially with T' set to 1, the generated image's garment depends heavily on the warped garment, which ensures ID preservation but can lead to distortions if the warped garment is distorted. Additionally, employing RePaint in full steps severely undermines the realism of the generated images. This is illustrated in Fig.~\ref{fig:repaint} last column, where there is a noticeable disconnection at the intersection of skirts and tops, and the shoulders are barely discernible. For the qualitative evaluation,  results across various metrics also indicate that setting $T' =1/2~T$, \ie, using RePaint for half of the steps during the inference, yields the best performance, as shown in Tab.~\ref{repaint_tab}.
}

\begin{table}[t!]
\centering
\caption{Quantitative comparison between the different methods of generating segmentation maps on the VITON-HD dataset~\cite{choi2021viton} at 512 $\times$ 384 resolution. $\textbf{MIoU}_{cloth}$ represents the result computed only within the clothing area, and $\textbf{MIoU}_{all}$ represents the result computed for the entire body excluding the neck. } 
    \setlength{\tabcolsep}{.4em}
\vspace{-5pt}
{\begin{tabular}{cccc}
\hline
    \textbf{Method} & {$\textbf{MIoU}_{cloth}\uparrow$}   & {$\textbf{MIoU}_{all}\uparrow$} \\ \hline
    VITON-HD~\cite{choi2021viton} &   0.8997  &   0.9598    \\    
    \textbf{Ours}	&\textbf{0.9662} & \textbf{0.9762}     \\ \hline
\end{tabular}}
\label{seg_compare:Tab}
\vspace{-10pt}
\end{table}

\begin{figure*} [h!]
	\centering
	\includegraphics[width=0.9\linewidth]{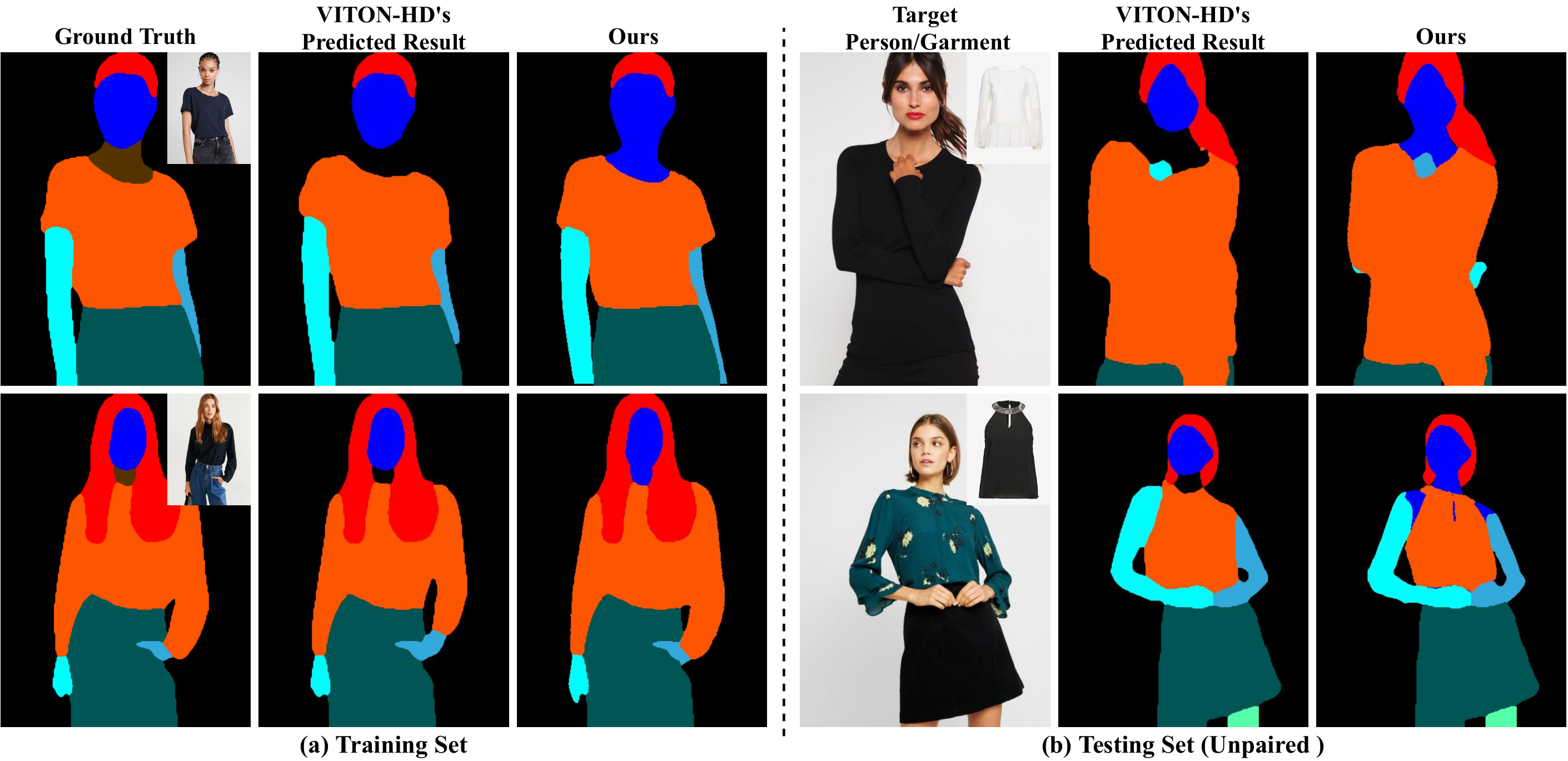}
 \vspace{-10pt}
	\caption{Qualitative Comparison between the different methods of generating segmentation maps on the VITON-HD dataset~\cite{choi2021viton} at 512 $\times$ 384 resolution. VITON-HD's predicted results come from a segmentation generator network, while our results are generated by a training-free method.}
	\label{fig:seg_compare}
 \vspace{-9pt}
\end{figure*}

\noindent \textbf{Comparison between the different methods of generating segmentation maps.} {To demonstrate the effectiveness of our designed training-free segmentation map generation method, we qualitatively and quantitatively compared our generated results with the results produced by VITON-HD~\cite{choi2021viton} using a trainable segmentation generator network. As for qualitative comparison, as the Training Set of the VITON-HD~\cite{choi2021viton} dataset provides the Ground Truth (GT) for the segmentation map, we calculated the \textbf{MIoU}~\cite{long2015fully} for the generated results of different methods against the GT. The calculation of \textbf{MIoU} includes $\textbf{MIoU}_{cloth}$ and $\textbf{MIoU}_{all}$, where the former computes only in the clothing area, while the latter computes in the entire body area. Since the segmentation generation network of VITON-HD~\cite{choi2021viton} does not generate the neck area, the calculation of $\textbf{MIoU}_{all}$ excludes the neck area. As shown in Tab.~\ref{seg_compare:Tab}, our method outperforms VITON-HD in both metrics, demonstrating the effectiveness of our approach. For the quantitative comparison, we conduct experiments on both the Training Set and Testing set (unpaired). As shown in Fig.~\ref{fig:seg_compare},  the results generated by both methods are very close to the Ground Truth on the Training Set. However, on the Testing Set, our method shows significantly better results. For example, in the first row, the result generated by VITON-HD is missing a hand, and in the second row, the generated cloth style is incorrect. Overall, our segmentation map generation method is very user-friendly, demonstrating good performance and requiring no network training parameters.
}

\begin{table*}[h!]
\centering
\caption{Quantitative comparison between Full Fine-tuning and Parameter Efficient Fine-tuning (PEFT) on the VITON-HD dataset~\cite{choi2021viton} at 512 $\times$ 384 resolution.}
\vspace{-10pt}
{\begin{tabular}{cccccccc}
\hline
    \textbf{Method} & {$\textbf{LPIPS}\downarrow$}   & {$\textbf{SSIM}\uparrow$} & {$\textbf{FID}_\textbf{u}\downarrow$}  & {$\textbf{KID}_\textbf{u}\downarrow$}  &\textbf{Params (Tunable)} & \textbf{Time (1 epoch)} \\ \hline
    Full Fine-tuning	&\textbf{0.068} &  \textbf{0.897}  & 8.63 & 0.79 & 1285M & 1.38 hour  \\ 
     \textbf{PEFT (Ours)} &   0.069  &  \textbf{0.897}  & \textbf{8.62}& \textbf{0.78} & \textbf{510M} & \textbf{0.83 hour}  \\  \hline
\end{tabular}}
\label{PEFT_tab}
\vspace{-10pt}
\normalsize
\end{table*}

\noindent \textbf{Comparison between Full Fine-tuning and Parameter Efficient Fine-tuning (PEFT).} {Our method is built upon Paint-by-Example~\cite{yang2023paint} and its pre-trained weights, thus inheriting the ability to manipulate specific areas while keeping others unchanged. Consequently, we only need to fine-tune the attention layers and the designed adapters that receive the critical identity cues to adapt to the try-on task. As shown in Tab.~\ref{PEFT_tab}, full fine-tuning only offers limited performance boosting on LPIPS and SSIM but leads to significant computational costs. We can also infer that the decoupled clothing identity, in conjunction with the injection modules we designed, have reduced the training difficulty and requirements of preserving the given garment. Therefore, considering the balance between performance and consumption, PEFT emerges as the preferred option.}


%% file: sec/5_conclusion.tex
\vspace{-5pt}
\section{Conclusion}

 Virtual try-on has gained widespread attention due to significantly enhancing the online shopping experience for users. We revisit {two} critical aspects of diffusion-based virtual try-on technology: identity controllability, and training efficiency. We propose an effective and efficient framework, termed TryOn-Adapter, to tackle these three issues. We first decouple clothing identity into fine-grained factors: style, texture, and structure. Then, each factor incorporates a customized lightweight module and fine-tuning mechanism to achieve precise and efficient identity control. {Meanwhile, we introduce a training-free technique, T-RePaint, to further reinforce the clothing identity preservation without compromising the overall image fidelity during the inference. In the final image try-on synthesis stage, we design an enhanced latent blending module for image reconstruction in latent space, enabling the consistent visual quality of the generated image.} Extensive experiments on two widely used datasets have shown that our method can achieve outstanding performance with minor trainable parameters. 

\noindent\textbf{Limitations.} {Although we satisfactorily resolve the issues of efficiently preserving the identity of the given garment and maintaining consistent visual quality for final try-on synthesis.  However, like most previous works, our method is still a certain distance away from achieving widespread practical application due to the limitation of the datasets. Furthermore, there is a lack of targeted quantitative evaluation metrics for virtual try-on tasks. We plan to develop a more granular evaluation from overall style, local texture, and structure for virtual try-on assessment, but progress is slow due to data scarcity.}

\noindent\textbf{Data Availability Statements.}
We claim to release the dataset and code upon acceptance. The datasets generated and analyzed during the current study will be available in our open-source repository.